# ProtTrans: Towards Cracking the Language of Life's Code Through Self-Supervised Learning

Ahmed Elnaggar, Michael Heinzinger, Christian Dallago,
Ghalia Rehawi, Yu Wang, Llion Jones, Tom Gibbs, Tamas Feher, Christoph Angerer,
Martin Steinegger, Debsindhu Bhowmik and Burkhard Rost

**Abstract**—Computational biology and bioinformatics provide vast data gold-mines from protein sequences, ideal for Language Models taken from NLP. These LMs reach for new prediction frontiers at low inference costs. Here, we trained two auto-regressive models (Transformer-XL, XLNet) and four auto-encoder models (BERT, Albert, Electra, T5) on data from UniRef and BFD containing up to 393 billion amino acids. The LMs were trained on the Summit supercomputer using 5616 GPUs and TPU Pod up-to 1024 cores. Dimensionality reduction revealed that the raw protein LM-*embeddings* from unlabeled data captured some biophysical features of protein sequences. We validated the advantage of using the *embeddings* as exclusive input for several subsequent tasks. The first was a per-residue prediction of protein secondary structure (3-state accuracy Q3=81%-87%); the second were per-protein predictions of protein sub-cellular localization (ten-state accuracy: Q10=81%) and membrane vs. water-soluble (2-state accuracy Q2=91%). For the per-residue predictions the transfer of the most informative embeddings (ProtT5) for the first time outperformed the state-of-the-art without using evolutionary information thereby bypassing expensive database searches. Taken together, the results implied that protein LMs learned some of the *grammar* of the *language of life*. To facilitate future work, we released our models at https://github.com/agemagician/ProtTrans.

---

## 1 INTRODUCTION

D EEP LEARNING (DL) has recently been advancing hand-in-hand with *High–Performance Computing* (HPC) to achieve new scientific breakthroughs in both fields. More powerful supercomputers [1], [2] and advanced libraries [3], [4], [5], [6], [7] enable the training of more complex models on bigger data sets using advanced processing units (incl. Graphics Processing Units (GPUs) and Tensor Processing Units (TPUs)).

Through contextualized Language Models (LMs) [8], [9], Natural Language Processing (NLP) has been benefiting substantially from advances in HPC. In particular *Transformers* [10] have reached state-of-the-art (*SOA*) performance for several tasks [11], [12]. Limitations in annotations do not constrain LMs: the self-supervised training exclusively relies upon the sequential order of the input, e.g., by reconstructing corrupted tokens given the surrounding sequence. After training, we can extract some information learned by the LMs, referred to as *embeddings*. *Transfer-learning* refers to the idea of using such embeddings as input for subsequently trained supervised models. These two steps outsource the computationally demanding LM pre-training to the HPC infrastructure, leaving the computationally less demanding inference to commodity hardware.

Proteins are the machinery of life, built from 20 different basic chemical building blocks (called *amino acids*). Like beads, those amino acids are strung up in one-dimensional (**1D**) sequences (the beads are referred to as *residues* once connected). These 1D sequences adopt unique three-dimensional (**3D**) shapes (referred to as protein *3D structure*) [13], and the 3D structures perform specific function(s) (often simplified as *sequence determines structure determines function*). We know many orders of magnitude more protein amino acid sequences than experimental protein structures (*sequence–structure gap*) [14]. Knowing protein structure helps to understand function. Closing, more generally, the sequence–annotation gap through prediction methods based on artificial intelligence (AI) is one of the crucial challenges for computational biology and bioinformatics. Tapping into the vast wealth of unlabeled data through transfer-learning may become crucial to bridging these gaps.

Top prediction methods in computational biology [15], [16], [17], [18], [19], [20] combine machine learning (<u>ML</u>) and *evolutionary information* (<u>EI</u>), first established as the winning strategy to predict protein secondary structure [21], [22] in two steps. First, search for a family of related proteins summarized as multiple sequence alignment (MSA) and extract the evolutionary information contained in this alignment. Second, feed the EI into the ML through supervised learning implicit structural or functional constraints. When predicting for proteins without experimental annotations, such methods only use ex-

- A. Elnaggar, M. Heinzinger, C. Dallago, G. Rehawi and B. Rost affiliated with TUM (Technical University of Munich) Department of Informatics, Bioinformatics & Computational Biology - i12, Boltzmannstr. 3, 85748 Garching/Munich, Germany.
- Y. Wang affiliated with Med AI Technology (Wu Xi) Ltd. , Ma Shan, Mei Liang Road, 88, 2nd floor (west), Bin Hu District, Wu Xi, Jiang Su Province, China.
- L. Jones affiliated with Google AI, Google, 1600 Amphitheatre Parkway, Mountain View, CA 94043, USA.
- T. Gibbs, T. Feher and C. Angerer affiliated with NVIDIA, 2788 San Tomas Expy, Santa Clara, CA 95051, Vereinigte Staaten, USA.
- M. Steinegger affiliated with School of Biological Sciences, Seoul National University, Seoul, 08826, South Korea.
- D. Bhowmik affiliated with Oak Ridge National Laboratory (ORNL), 1 Bethel Valley Rd, Oak Ridge, TN 37830, Vereinigte Staaten.
- A. Elnaggar & M. Heinzinger contributed equally to this work.
- Corresponding author: ahmed.elnaggar [at] tum.de, tel: +49-289-17-811 (email rost: assistant [@] rostlab.org)
- The official GitHub repository: https://github.com/agemagician/ProtTrans

perimental information implicitly captured in the trained model. Since all other information originates from the knowledge of sequences, such methods need no additional information as input other than the EI which is amply available giving the exploding databases of bio-sequences [23], [24]. However, there are several prices to pay for EI. Firstly, when predicting for entire proteomes (all proteins in an organism), compiling the EI for all is computationally expensive [25]. Secondly, EI is not available for all proteins (intrinsically disordered proteins [26] or *dark proteome* [27]). Thirdly, the improvement is best when the EI is most diverse [28], [29]. Fourthly, predictions based on EI somehow average over an entire family, possibly falling short of distinguishing differences between two different proteins in the same family. The latest, and arguably largest leaps ever in terms of protein structure prediction, namely AlphaFold2, roots on an advanced combination of EI and ML [30]. Although that method predicts protein 3D structures at unprecedented levels of accuracy, AlphaFold2 models are many order of magnitude more computationally expensive than the creation of EI.

The leap of NLP through advanced LMs has been successfully generalized toward understanding the *language of life* through advanced LMs trained on proteins [31], [32], [33], [34], [35], [36], [37], [38], [39]. In analogy to NLP, these approaches interpret an entire protein sequence as a sentence and its constituents – amino acids – as single words. Protein sequences are constrained to adopt particular 3D structures optimized for accomplishing particular functions. These constraints mirror the rules of grammar and meaning in NLP. Since LMs extract features directly from single protein sequences, they might reach performance of the SOA without using EI.

In this project, dubbed *ProtTrans*, we pursued two objectives. Firstly, we explored the limits of up-scaling language models trained on proteins as well as protein sequence databases used for training. Secondly, we compared the effects of autoregressive and auto-encoding pre-training upon the success of the subsequent supervised training, and compared all LMs trained here to existing state-of-the-art (SOA) solutions using evolutionary information (EI) [40].

## 2 METHODS

### 2.1 Data for protein Language Models (LMs)

In this work, we assessed the impact of database size on performance through three data sets (Table 1, SOM Fig. 10 ): Uniref50 [41], UniRef100 [41], and BFD [24], [42]. The latter merged UniProt [23] and proteins translated from multiple metagenomic sequencing projects, making it the largest collection of protein sequences available at the time of writing even after removal of duplicates from the original BFD. Overall, BFD was about eight times larger than the largest data sets used previously for protein LMs [34]. Despite the 8-fold increase in data, the number of tokens increased only five-fold (Table 1), because UniRef100 sequences were longer than those in BFD (1.6-fold). Without a clear mapping for LMs from NLP to proteins, i.e., the concept of words can be related to single amino acids, a window of amino acids (k-mer motifs [43]) or functional units (domains [44]), we decided to interpret single amino acids as input tokens/words. Thereby, protein databases contain several orders of magnitude more tokens than corpora used in NLP, e.g., Google's Billion Word data set [45] is one of the biggest for NLP with about 829 million tokens (words), i.e. about 500-times fewer than BFD with 393 billion tokens. Interpreting domains as words, would cut the number of tokens in BFD roughly by a factor of 100 (average domain length [46]) still leaving 5-times more tokens in BFD than the Billion Word corpus. Uniref50, UniRef100 and BFD were tokenized with a single space (indicating word-boundaries) between each token. Each protein sequence was stored on a separate line, with lines/proteins representing the equivalent of "sentences". Additionally, an empty line was inserted between each protein sequence in order to indicate the "end of a document"; however, this is only essential for models with auxiliary task (Bert and Albert). Non-generic or unresolved amino acids ([BOUZ]) were mapped to *unknown* (X). For training ProtTXL and ProtT5, the data was transformed to pytorch and tensorflow tensors, respectively on the fly. For ProtBert, ProtAlbert, ProtXLNet and ProtElectra, the data was pre-processed and stored as tensorflow records. Given tensorflow records with terabytes, data sets had to be chunked into 6000 files for thousands of parallel workers.

| Data LM | UniRef50 | UniRef100 | BFD |
|---|---|---|---|
| *Number proteins* [in m] | 45 | 216 | 2,122 |
| *Number of amino acids* [in b] | 14 | 88 | 393 |
| *Disk space* [in GB] | 26 | 150 | 572 |

TABLE 1: Data Protein LM - UniRef50 and UniRef100 cluster the UniProt database at 50% and 100% pairwise sequence identity (100% implying that duplicates are removed) [41]; BFD combines UniProt with metagenomic data keeping only one copy for duplicates [24], [42]. Units: number of proteins in millions (m), of amino acids in billions (b), and of disk space in GB (uncompressed storage as text).

### 2.2 *Embeddings* for supervised training

We extracted the information learned by the protein LMs through *embeddings*, i.e., vector representations from the last hidden state of the protein LM (Fig. 1). In the transfer-learning step these embeddings served as input to subsequent supervised training. Although we mostly relied on previously published data sets to ease comparisons to other methods, for the supervised training, we also added a novel test set to refine the evaluation.

**Per-residue prediction/single tokens:** to predict properties of single tokens (here: single amino acids, dubbed residues when joined in proteins), we used the training set published with NetSurfP-2.0 [15] describing secondary structure in 3- and 8-states (class distribution for all data sets in SOM Tables 6, 5). We also included other public test data sets, namely CB513 [47], TS115 [48], and CASP12 [49]. Each of those has severe limitations (CASP12: too small, CB513 and TS115 redundant and outdated). Therefore, we added a new test set using only proteins published after the release of NetSurfP-2.0 (after Jan 1, 2019). We included proteins from the PDB [50] with resolutions ≤ 2.5Å and ≥ 20 residues. MMSeqs2 [51] with highest sensitivity (-s 7.5) removed proteins with >20% PIDE to either the training set or to itself. On top, PISCES [52] removed any protein constrained by its procedure to have >20% PIDE. These filters reduced the number of new proteins (chains) from 18k to 364 (dubbed set *NEW364*).

**Per-protein prediction/embedding pooling:** For the prediction of features for entire proteins (analogous to the classification of whole sentences in NLP), the DeepLoc [16] data set was used to classify proteins into (i) membrane-bound vs. water-soluble and (ii) ten classes of subcellular localization (also referred to as cellular compartments).

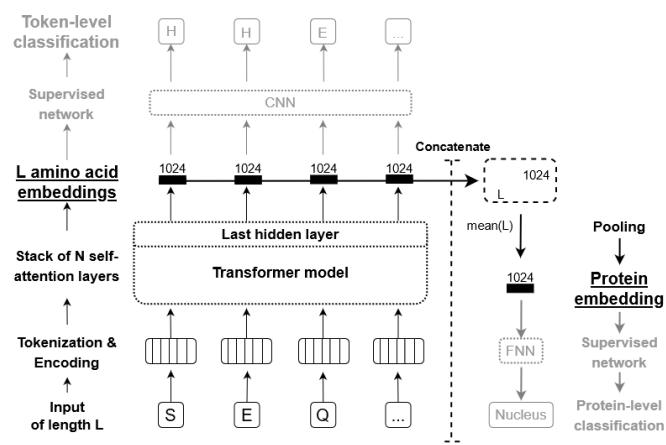

Fig. 1: Feature extraction overview - We give a general overview on how ProtTrans models can be used to derive features (embeddings) for arbitrary protein sequences either on the level of single amino acids or whole proteins and how they can be used for classification tasks on both levels. First, an example protein sequence "SEQ" is tokenized and positional encoding is added. The resulting vectors are passed through any of our ProtTrans models to create context-aware embeddings for each input token, i.e. each amino acid. Here, we used the last hidden state of the Transformer's attention stack as input for downstream prediction methods. Those embeddings can either be used directly as input for prediction tasks on the level of individual tokens, e.g. a CNN can be used to predict an amino acid's secondary structure. Alternatively, those embeddings can be concatenated and pooled along the length-dimension to get fixed-size embedding irrespective of the input length, i.e., global average pooling is applied. The resulting protein-level embedding can be used as input for predicting aspects of proteins, e.g., a FNN can be used to predict a protein's cellular localization.

## 2.3 Data for unsupervised evaluation of embeddings

We also assessed the information captured by the embeddings extracted from the protein LMs by projecting the high-dimensional representations down to two dimensions (2D) using t-SNE [53]. Toward this end, we took annotations from several sources. First, a non-redundant (PIDE<40%) version of the SCOPe database [54] (release 2.07 with 14,323 proteins). Second, we mapped proteins into the three major domains of life (archaea, bacteria, or eukarya) or to viruses (removing all proteins with missing classifications). The number of iterations for the t-SNE projections was set to 3,000 and the perplexity to 30 for all plots with the exception of the amino acid plot for which we used a perplexity of 5. All visualizations used the same random seed (42).

## 2.4 Step 1: Protein LMs extract embeddings

We trained six successful LMs in NLP (T5 [55], Electra [56], BERT [57], Albert [58], Transformer-XL [59] and XLNet [11]) on protein sequences. *BERT* was the first bidirectional model in NLP which tried to reconstruct corrupted tokens, and is considered the de-facto standard for transfer learning in NLP. *Albert* reduced BERT's complexity by hard parameter sharing between its attention layers which allows to increase the number of attention heads (64 chosen here). *Electra* tries to improve the sampling-efficiency of the pre-training task by training two networks, a generator and a discriminator. Instead of only reconstructing corrupted input tokens, the generator (BERT) reconstructs masked tokens, potentially creating plausible alternatives, and the discriminator (Electra) detects which tokens were masked. This enriches the training signal as the loss can be computed over all tokens instead of the subset of corrupted tokens (usually only 15%). *T5* uses the original transformer architecture proposed for sequence translation, which consists of an encoder that projects a source language to an embedding space and a decoder that generates a translation to a target language based on the encoder's embedding. Only later, models used either the encoder (BERT, Albert, Electra) or the decoder (*TransformerXL, XLNet*), but T5 showed that this simplification might come at a certain prize as it reaches state-of-the-art results in multiple NLP benchmarks. Additionally, it provides the flexibility to apply different training methodologies and different masking strategies, e.g., T5 allows to reconstruct spans of tokens instead of single tokens.

As self-attention is a set-operation and thus order-independent, Transformers require explicit positional encoding. Models trained with sinusoidal position signal like BERT, Albert or Electra, can process only sequences shorter or equal to the length of the positional encoding which has to be set before training. Due to the huge memory requirement of Transformer-models, this parameter is usually set to a value lower than the longest proteins, e.g., Titin with 33k residues. Here, we trained models that were affected by this limitations (ProtBERT, ProtAlbert, ProtElectra) first on proteins of length ≤ 512, then on proteins ≤ 1024. Only setting the length of the positional encoding to 40k after pre-training allowed the models to process protein sequences up to a length of 40k. In contrast to this, TransformerXL introduced a memory that allows it to process sequences of arbitrary length. Still, the model cuts sequences into fragments but allows for flow of information between fragments for longer proteins by re-using hidden states of fragments which have already been processed. While its memory is uni-directional as fragments are processed sequentially, TransformerXL captures only uni-directional context within one memory fragment (auto-regressive) while XLNet, which uses a similar memory mechanism to process sequences of arbitrary length, allows to gather bidirectional context within one memory fragment.

In contrast to this, T5 learns a positional encoding for each attention head that is shared across all layers. This way, the model learned to combine the relative offset between residue pairs of lower layers, enabling the model to make predictions beyond the actual length of the positional encoding. No auxiliary tasks like BERT's next-sentence prediction were used for any model described here.

ProtTXL, ProtBert, ProtXLNet, ProtAlbert and ProtElectra were trained on UniRef100, ProtT5 on UniRef50, and ProtTXL, ProtBert & ProtT5, on BFD (Table 2). Largely, we transferred configurations successfully from NLP to protein sequences [36], [39], [60], with the exception of the number of layers that was increased to optimize memory utilization.

**ProtTXL:** The Transformer-XL[1] was trained using both UniRef100 and BFD-100 datasets (referred to as *ProtTXL* and *ProtTXL-BFD*, respectively; Table 2). Both models used a dropout rate of 15%, a memory length of 512 tokens and using mixed precision. The number of layers, number of heads, batch size, learning rate, weight decay, training steps and warm-up steps were adjusted according to training set size as well as GPU utilization. The number of warm-up steps was set to cover at least one epoch for each data set. We tested initial learning rates between 0.001 and 0.005 which were increased linearly at every training step over the warm-up period. To avoid model divergence during training, the learning rate had to be (i) reduced along with the warm-up steps (for BFD), or (ii) increased for both (for Uniref100). Even after increasing the warm-up steps to two epochs, the maximum learning rate remained at 0.0025 for both data sets. Beyond this point, the training diverged. Using weight decay to regularize the network increased the GPU memory usage as it required to compute the norm of all weight vectors on our models, thus reducing the batch size. ProtTXL-BFD was trained for 40k steps in total, with 13.6k warm-up steps using a learning rate of 0.0005, while ProtTXL was trained for 31k steps with 5k warm-up steps using a learning rate of 0.002. The Lamb optimizer was able to handle the resulting batch sizes of 44k and 22k for ProtTXL-BFD and ProtTXL, respectively, without divergence.

---

1. https://github.com/NVIDIA/DeepLearningExamples/

| Hyperparameter | ProtTXL | | ProtBert | | ProtXLNet | ProtAlbert | ProtElectra | ProtT5-XL | | ProtT5-XXL | |
|---|---|---|---|---|---|---|---|---|---|---|---|
| Dataset | BFD100 | UniRef100 | BFD100 | UniRef100 | UniRef100 | UniRef100 | UniRef100 | UniRef50 | BFD100 | UniRef50 | BFD100 |
| Number of Layers | 32 | 30 | 30 | 30 | 30 | 12 | 30 | 24 | 24 | 24 | 24 |
| Hidden Layers Size | 1024 | 1024 | 1024 | 1024 | 1024 | 4096 | 1024 | 1024 | 1024 | 1024 | 1024 |
| Hidden Layers Intermediate Size | 4096 | 4096 | 4096 | 4096 | 4096 | 16384 | 4096 | 16384 | 16384 | 65536 | 65536 |
| Number of Heads | 14 | 16 | 16 | 16 | 16 | 64 | 16 | 32 | 32 | 128 | 128 |
| Positional Encoding Limits | - | - | 40K | 40K | - | 40K | 40K | - | - | - | - |
| Dropout | 0.15 | 0.15 | 0.0 | 0.0 | 0.1 | 0.0 | 0.0 | 0.1 | 0.1 | 0.1 | 0.0 |
| Target Length | 512 | 512 | 512/2048 | 512/2048 | 512 | 512/2048 | 512/1024 | 512 | 512 | 512 | 512 |
| Memory Length | 512 | 512 | - | - | 384 | - | - | - | - | - | - |
| Masking Probability | - | - | 15% | 15% | | 15% | 25% | 15% | 15% | 15% | 15% |
| Local Batch Size | 8 | 5 | 32/6 | 30/5 | 2 | 21/2 | 18/7 | 8 | 4 | 8 | 4 |
| Global Batch Size | 44928 | 22464 | 32768/6144 | 15360/2560 | 1024 | 10752/1024 | 9216/3584 | 2048 | 4096 | 2048 | 4096 |
| Optimizer | Lamb | Lamb | Lamb | Lamb | Adam | Lamb | Lamb | AdaFactor | AdaFactor | AdaFactor | AdaFactor |
| Learning Rate | 0.0005 | 0.002 | 0.002 | 0.002 | 0.00001 | 0.002 | 0.002 | 0.01 | 0.01 | 0.01 | 0.01 |
| Weight Decay | 0.0 | 0.01 | 0.01 | 0.01 | 0.01 | 0.01 | 0.01 | 0.0 | 0.0 | 0.0 | 0.0 |
| Training Steps | 40.7K | 31.3K | 800K/200K | 300K/100K | 847K | 150K/150K | 400K/400K | 991K | 1.2M | 343K | 920K |
| Warm-up Steps | 13.6K | 5.5K | 140K/20K | 40K/0K | 20K | 40K/5K | 40K/40K | 10K | 10K | 10K | 10K |
| Mixed Precision | FP16 Model Weight Fp32 Master Weight | | None | None | None | None | None | None | None | None | None |
| Number of Parameters | 562M | 409M | 420M | 420M | 409M | 224M | 420M | 3B | 3B | 11B | 11B |
| System | Summit | Summit | TPU Pod | TPU Pod | TPU Pod | TPU Pod | TPU Pod | TPU Pod | TPU Pod | TPU Pod | TPU Pod |
| Number of Nodes | 936 | 936 | 128 | 64 | 64 | 64 | 64 | 32 | 128 | 32 | 128 |
| Number of GPUs/TPUs | 5616 | 5616 | 1024 | 512 | 512 | 512 | 512 | 256 | 1024 | 256 | 1024 |

TABLE 2: Large-scale Deep Learning: the table shows the configurations for pre-training the protein LMs introduced here (ProtTXL, ProtBert, ProtXLNet, ProtAlbert, ProtElectra, ProtT5) using either Summit, a TPU Pod v2 or a TPU Pod v3.

**ProtBert:** BERT[2] was trained using both UniRef100 and BFD-100 datasets (referred to as *ProtBert* and *ProtBert-BFD*, respectively; Table 2). Compared to the original BERT publication, the number of layers was increased. Unlike Transformer-XL which was trained on Nvidia GPUs, mixed-precision was not used to train other models because those were trained on TPUs. Similar to the BERT version trained in the Lamb paper [61], ProtBert was first trained for 300k steps on sequences with a maximum length of 512 and then for another 100k steps on sequences with a length of a maximum length of 2k. While ProtBert-BFD was trained for 800k steps, then for another 200k steps for sequences with maximum length of 512 and 2k, respectively. This allows the model to first extract useful features from shorter sequences while using a larger batch size, rendering training on longer sequences more efficient.

**ProtAlbert:** We trained Albert[3] on UniRef100 (*ProtAlbert*; Table 2). We used the configuration from the official GitHub

repository for Albert (version: xxlarge v2). For Albert the number of layers is increased through the number of times, Albert stacks its single layer. Compared to the original publication, we achieved increasing the global batch size from 4096 to 10752 on the same hardware. The reason for this counter-intuitive effect is the reduced vocabulary size in proteins: the entire diversity of the protein universe is realized by 20 different amino acids, compared to tens of thousands of different words. Similar to ProtBert, ProtAlbert was first trained for 150k steps on sequences with a maximum length of 512 and then for another 150k steps on sequences with a maximum length of 2k.

**ProtXLNet:** XLNet[4] was trained on UniRef100 (*ProtXL-Net*) using the original NLP configuration [11] (Table 2) except for the number of layers that was increased to 30 layers which reduced the global batch size to 1024. Due to the relatively small batch-size, we used the original optimizer: Adam with a learning rate of 0.00001. The model was trained through more steps, i.e. 20k warm-up and 847k steps to compensate for the smaller batch-size of this model.

**ProtElectra:** Electra[5] consists of two models, a generator and discriminator (same number of layers, generator 25% of the discriminator's hidden layer size, hidden layer intermediate size, and number of heads). We copied Electra's NLP configuration with two changes: increasing the number of layers to 30 and using Lamb optimizer. Again, we split the training into two phases: the first for proteins ≤ 512 residues (400k steps at 9k global batch size), the second for proteins ≤ 1024 (400k steps at 3.5k global batch size). While ProtTXL, ProtBert, ProtAlbert and ProtXLNet relied on pre-computed tensorflow records as input, Electra allowed to mask sequences on the fly, allowing the model to see different masking patterns during each epoch.

**ProtT5:** Unlike the previous LMs, T5[6] uses an encoder and decoder [10]. We trained two model sizes, one with 3B (T5-XL) and one with 11B parameters (T5-XXL). T5-XL was trained using 8-way model parallelism, while T5-XXL was trained using 32-way model parallelism. First, T5-XL and T5-

---

XXL were trained on BFD for 1.2M and 920k steps respectively (*ProtT5-XL-BFD, ProtT5-XXL-BFD*). In a second step, ProtT5-XL-BFD and ProtT5-XXL-BFD were fine-tuned on UniRef50 for 991k and 343k steps respectively (*ProtT5-XL-U50, ProtT5-XXL-U50*). Contrary to the original T5 model which masks spans of multiple tokens, we adopted BERT's denoising objective to corrupt and reconstruct single tokens using a masking probability of 15%. All T5 models used the AdaFactor optimizer with inverse square root learning rate schedule for pre-training. Like ProtElectra, T5 masks each sequence on the fly. In our hands, the encoder outperformed the decoder on all benchmarks significantly and running the model in half-precision during inference instead of full-precision had no effect on performance but allowed to run the model on on a single Nvidia TitanV (12GB vRAM). Thus, we dropped the decoder from further analysis which cuts model size by half during inference. For completeness, we made weights for encoder and decoder publicly available.

### 2.5 Step 2: Transfer learning of supervised models

To best analyze the impact of transfer learning, we deliberately kept the supervised models using the embeddings from the protein LMs as input minimal. In particular, compared to SOA solutions such as NetSurfP-2.0, all our experiments used the pre-trained LMs as feature extractors without fine-tuning, i.e. without gradient back-propagating to the LMs. Throughout, we extracted the embeddings from the last hidden state of the pre-trained LMs as described in detail elsewhere [32]. To briefly summarize (Fig. 1): we applied tasks on two different levels, namely on the level of single tokens (per-residue) and whole sentences through pooling (per-protein) predictions. For the **per-residue prediction**, we input the embeddings into a two-layer convolutional neural network (CNN). The first CCN layer compressed the embeddings to 32 dimensions using a window size of 7. The compressed representation was fed into two different CNNs (each with window size 7). One learned to predict secondary structure in 3-states, the other in 8-states. The network was trained on both outputs simultaneously by adding their losses (multi-task learning). For ProtBERT-BFD embeddings we additionally trained three other models: logistic regression, FNN and LSTM. Similar to the CNN, the two-layer FNN first compressed the output of the language model down to 32 dimensions which the second FNN-layer used to predict 3- and 8-states simultaneously. The bi-directional LSTM compressed the embeddings down to 16 dimensions. Concatenating both directions, the resulting 32 dimensional representation was used by a FNN layer to predict 3- or 8-states. As the CNN performed best (SOM Table 10), we used CNNs throughout. For the **per-protein prediction**, we also extracted the embeddings from the last layer of the protein LMs. However, then we pooled the representations over the length-dimension resulting in a fixed-size representation for all proteins. Using ProtBERT-BFD embeddings, we compared alternative pooling strategies (SOM Table 10) and chose mean-pooling for all further experiments. The resulting vector was used as an input to a single feed forward layer with 32 neurons which compressed information before making the final predictions for both per-protein tasks, i.e., the prediction of subcellular localization and the differentiation between membrane-bound and water-soluble proteins, simultaneously (multi-task learning).

### 2.6 Hardware

HPC hardware is advancing both through infrastructure of supercomputers, such as Fugaku [62], Summit [1] or the SuperMUC-NG [63], and through its components, such as TPU pods [2], specifically designed to ease large scale neural network training for users. Concurrent software improvements in form of more efficient libraries such as Horovod [6] allow executing general purpose code on large distributed clusters with minor code changes. In this section we give details on the hard- and software used for training language models on large protein sequence databases.

**ORNL Summit & Rhea:** The Oak Ridge National Laboratory (ORNL) provides several clusters for researchers who need computational resources not provided by research facilities such as universities. Here, we used *Summit* and *Rhea*. Summit was used to train the deep learning models, while Rhea was used for the pre-processing of data sets including the distributed generation of tensorflow records.

Summit is the world's second fastest computer, consisting of approximately 4618 nodes. Each node has two IBM POWER9 processors and six NVIDIA Volta V100 with 16GB of memory each [1]. Every POWER9 processor is connected via dual NVLINK bricks, each capable of a 25GB/s transfer rate in both directions. A single node has 0.5 TB of DDR4 main memory and 1.6TB of non-volatile memory that can be used as a burst buffer. Summit is divided into racks with each rack having 18 nodes. In all of our experiments we reserved 936 nodes for training. As having nodes on the same rack decreases the communication overhead, we reserved entire racks.

The smaller cluster (Rhea) contains two partitions: Rhea and GPU. The Rhea partition has 512 node, each with 128 GB of memory and two Intel® Xeon® E5-2650. The GPU partition has only 9 nodes, each with 1 TB of memory and two Intel® Xeon® E5-2695. Reha reduced the time needed for creating tensorflow records for the BFD dataset from 7.5 months (!) to fewer than two days, by converting the original sequential script to distributed processing using MPI. The generation script used two nodes of the GPU partition, with a total of 112 parallel threads.

**Google TPU Pod:** In 2016, Google introduced tensor processing unit (TPU) as its application-specific integrated circuit optimized for training neural networks. TPUs can be accessed through Google Cloud. Training the protein LMs used both older TPU generation (V2) with 256 cores, and the latest TPU generation (V3) with 512 and 1024 cores. These cores are divided into hosts with each host having access to 8 cores. Consequently, we had access to 32, 64 and 128 hosts for V2/V3-256, V3-512 and V3-1024, and each core had 8 GiB and 16 GiB of high-bandwidth memory for V2 and V3. Training on the TPUs required access to a virtual machine on Google Cloud and storage on Google Bucket [64].

### 2.7 Software

Summit integrates several pre-configured modules which include the most popular libraries and tools required for simulation, deep learning, distributed training and other purposes. We used the IBM Watson Machine Learning module versions 1.6.0 and 1.6.2 for our deep learning training. In contrast to this, the Google Cloud server, which we used for the TPU Pod training, had to be configured manually because only the operating system was installed.

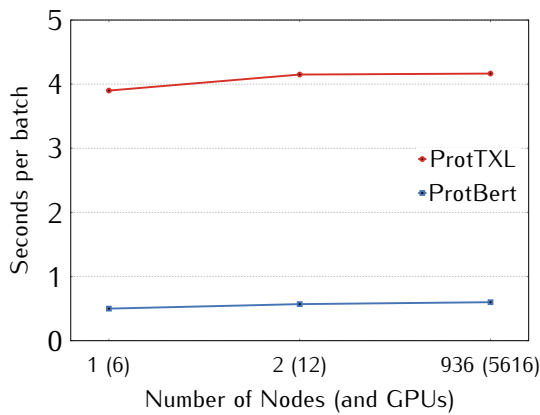

Fig. 2: Large Scale Dataset Training: The figure shows the overhead of increasing the number of nodes/gpus for both ProtTXL (blue; low) and ProtBert (red; high). The overhead increases slightly from 1 to 2 nodes but remains constant even when scaling up to 936 nodes with a total of 5616 GPUs. Having a low overhead means the model has a near-linear scale-up across thousands of GPUs, upper-bounded by the theoretical scale-up.

Pytorch was used to train ProtTXL, tensorflow to train ProtBert, ProtAlbert, ProtXLNet, ProtElectra and ProtT5. Both libraries used the Horovod framework [6] to train the models on distributed clusters such as Summit. Horovod supports distributed GPU training with minimal change in the code. It supports different backends including MPI, NCCL and IBM PowerAI distributed deep learning (DDL). We tested all three backends and found DDL to be the fastest for our training purpose on Summit. The time needed to finish a single batch with ProtTXL-BFD increased from one to two nodes due to the communication overhead (Fig. 2). After two nodes the communication overhead plateaued, even when scaling up to 936 nodes with 5616 GPUs. Summit has integrated DDL in their Watson Machine Learning module which comes with most DDL libraries including pytorch, tensorflow, apex, DDL and horovod. However, Summit has only a license for using DDL up to 954 nodes. Contrary to Summit, training on TPU Pods did not require any changes in the Tensorflow code to use either a single TPU host or to distribute workload among multiple TPU hosts.

Mixed precision allows to fit larger models and batch sizes into GPU memory by using 16-bit precision only or a mix of 16-bit and 32-bit precision. Nvidia's APEX library [65] was used for mixed precision training of ProtTXL, because APEX supports pytorch. APEX supports four types of mixed precision and model weights storing: 1) Pure 32-bit precision; this is the regular training without using mixed precision. 2) Pure 16-bit precision, all the model weights will be stored in 16-bit rather than 32-bit. 3) Mixed Precision, for different layer types depends on previously tested whitelist/blacklist by Nvidia; some weights will be stored in 32-bit while others in 16-bit format. 4) Almost FP16, storing all model weights at 16 Bit precision; exception: batch-normalization layers, while keeping a master copy of the model's weights in 32-Bit. Using pure 16-bit training leads to a big part of activation gradient values becoming zeros, leading to divergence during training. This problem is solved using Almost FP16 because there is a master copy of the model's weights in 32-Bit. As ProtTXL training became instable when training with 16 Bit precision, we switched to almost half precision training. We did not use mixed-precision for models trained on TPUs.

Another optimization technique/library crucial for our training on Summit was IBM's large model support (LMS) [66]. Similar to gradient checkpointing [67], LMS virtually extends the GPU memory by outsourcing parts of the model from GPU to main memory. This allows training models larger than the GPU memory. The obvious drawback of LMS is the increase in training time due to shuttling data between CPU and GPU and back. However, the reduced memory consumption of the model allows to increase the batch size, potentially compensating for the communication overhead. Compared to gradient checkpointing, LMS provides easier integration into existing code by operating directly on a computational graph defined by users and automatically adds swap-in and swap-out nodes for transferring tensors from GPU memory to main memory and vice versa. We have tested LMS on ProtTXL as well as ProtBert (Figure 2). As Pytorch and tensorflow have different strategies to integrate LMS, we also compared the effect of LMS on batch-size, model size and training time using the two different libraries. ProtTXL was used to evaluate the effect of Pytorch's implementation of LMS while ProtBert was trained for a few steps BFD using Summit to evaluate tensorflow's implementation of LMS. Training ProtBert for a few steps was sufficient to assess the effect of LMS on batch-size, model size as well as an estimate of training time. In the end, we used LMS only for ProtTXL to strike a balance between model size and training time. The number of LM parameters could be increased by about 15.6% for ProtTXL-BFD and to 6.6% for ProtBert (3a). Additionally, we could increase the batch size by 700% for ProtTXL-BFD (Figures 3b and 3c). The NV-Link between CPU and GPU on Summit-nodes, reduced the training time for ProtTXL by 60%while it increased by 72% for ProtBert (Figure 3d).

## 3 RESULTS

### 3.1 Step 1: Unsupervised protein LMs informative

Embeddings extract constraints about protein function and structure learned by the protein LMs in the first self-supervised step of pre-training on raw protein sequences. Using t-SNE [53], we visualized this information by projecting the embeddings onto 2D and annotated structural, functional or evolutionary features. Using attention maps, we analyzed the DNA-binding zinc-finger motif well conserved in evolution.

**Capturing biophysical features of amino acids.** Applying t-SNE to the uncontextualized token embedding layer visualized information extracted by the LMs for individual amino acids independent of their context (residues next to it). As previously established for another protein LM [39], the t-SNE projections (e.g. ProtT5-XL-U50 SOM Fig. 14A or ProtBert-BFD SOM Fig. 15A) suggested that all LMs captured essential biophysical amino acid features, including charge, polarity, size, hydrophobicity, even to the level of aliphatic ([AILMV]) vs. aromatic ([WFY]).

We compared the embedding projection with a randomly initialized model of identical architecture to ascertain that the observed effects did not originate from coincidental signals originating from projecting high-dimensional data (Fig. 4A) or some inductive bias of neural networks [68]. The random projection clearly did not carry biophysical information, while the embeddings projection did.

   

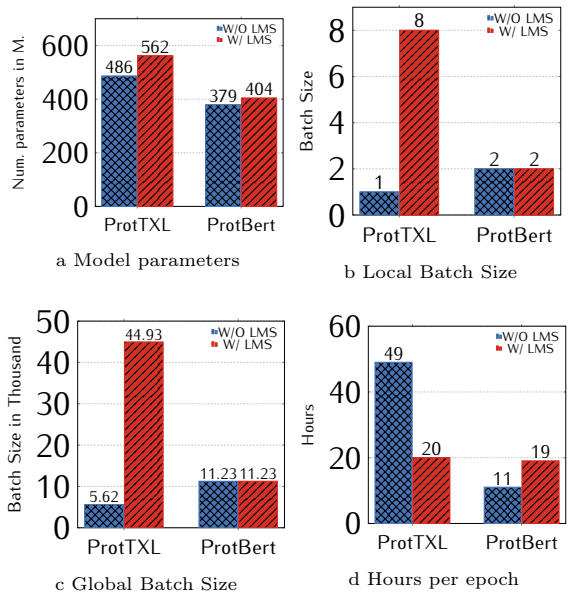

Fig. 3: Large Scale Deep Learning Training: The figures show the effect of enabling (red bars) or disabling (blue bars) large model support (LMS) on both, model size as well as batch size, when we tested ProtTXL or ProtBert on Nvidia V-100 16GB GPUs. It highlights the difference between applying LMS using PyTorch (ProtTXL) or tensorflow (ProtBert). Panel (a) shows the effect of using LMS on the maximum model size that can fit in a single V-100 memory. Panels (b,c) compare the effect of LMS on the maximum local (b) and global batch size (c) that can fit in the GPU. The number of hours required to finish a single epoch using 936 nodes, each with 6 GPUs when LMS being enabled is shown in (d).

**Capturing protein structure classes.** To assess which aspects of protein structure were captured by the unsupervised LMs, we averaged over the length-dimension of the representations derived from the last layer of each model (see Fig. 1 for a sketch) to derive fixed size representations for each protein in the database. We annotated structural class through the SCOPe database [54] (Methods). ProtBert-BFD and especially ProtT5-XL-U50 embeddings visually separated the proteins best (details in SOM Figs. 14-19). Although sequence length was not explicitly encoded and our pooling squeezed proteins into a fixed vector size, all models separated small from long proteins (brown, e.g. ProtT5-XL-U50 SOM Fig. 14D). All models also distinguished between water-soluble and transmembrane proteins (light blue, e.g. ProtT5-XL-U50 SOM Fig. 14D) and, to some extent, between proteins according to their secondary structure composition(e.g. all-alpha (dark blue) vs. all-beta (dark green) ProtT5-XL-U50 Fig. 14D). While having much higher entropy, even the random clustered small proteins from long proteins (brown, Fig. 4B).

**Capturing domains of life and viruses.** The analysis distinguished three domains of life: *archaea*, *bacteria*, and *eukarya*, along with *viruses* typically not considered as life. We used the same proteins and per-protein pooling as for the SCOPe analysis. All protein LMs captured some organism-specific aspects (e.g. ProtT5-XL-U50 SOM Fig. 14E). Eukarya and bacteria clustered better than viruses and archaea. Comparing different LMs revealed the same trend as for protein structure classes: ProtTXL (SOM 19E) and ProtBert (SOM 16E) produced higher entropy clusters while ProtAlbert (SOM 17E), ProtXLNet (SOM 18E), ProtBERT-BFD (SOM Fig. 15E) and ProtT5-XL-U50 (SOM Fig. 14E) produce visually easier separable clusters.

**Capturing protein function in conserved motifs.** A similar overall per-protein analysis as for structural classes and domains of life also suggested some clustering according to protein function as proxied by enzymatic activity (EC-numbers [69] and subcellular localization (SOM –1.2 Protein LMs unsupervised). We focused in more detail on the attention mechanism [70] at the core of each Transformer model [10] providing some limited understanding of the AI [71], [72]. We visualized [73] the attention weights of ProtAlbert to analyze the structural motif of a zinc-finger binding domain (SOM Fig. 11) crucial for DNA- and RNA-binding and conserved across diverse organisms. The right part of ProtAlbert' attention heads (SOM Fig. 11; line thickness resembles attention weight) learned to focus mostly on the four residues involved in zinc-binding (residues highlighted in the left part of SOM Fig. 11) which is essential for function.

### 3.2 Step 2: Embeddings good input to predict

The acid test for proving that the embeddings from protein LMs extracted important constraints is to exclusively use embeddings as input to supervised training of features reflecting structure and function. We proxied this through predictions on two different level, namely on the <u>per-residue or token level</u> (secondary structure) and on the <u>per-protein or sentence level</u> through pooling over entire proteins (localization, and classification into membrane/non-membrane proteins). Protein LMs remained unchanged, i.e. both approaches (per-residue/per-protein) used only embeddings derived from the hidden state of the last attention layer of each protein LM (Fig. 1) without gradient backpropagation to the LM, i.e., LMs were only used as static feature extractors.

#### 3.2.1 Per-residue secondary structure prediction

To ease comparability, we evaluated all models on standard performance measures (Q3/Q8: three/eight-state per-residue accuracy, i.e. percentage of residues predicted correctly in either of the 3/8 secondary structure states) and on standard data sets (CASP12, TS115, CB513). To increase the validity of the comparisons, we added a novel, non-redundant test set (dubbed *NEW364*, Methods). For simplicity, we only presented values for Q3 on CASP12 and NEW364 (TS115 and CB513 contain substantial redundancy; Q8 results brought little novelty; all details in SOM Tables 9, 8). As error estimates failed to capture the performance variation between NEW364 and CASP12, we used CASP12 as lower- and NEW364 as upper-limit.

*Comparing supervised architectures:* We input embeddings from ProtBERT-BFD into four different models for supervised training (Methods): logistic regression (LogReg), FNN, CNN and LSTM. LogReg provided an advanced *baseline* (Q3(LogReg)=74.3-79.3, where the spread is from the lower level for the set CASP12 and the upper for the set NEW364; SOM Table 7). LSTMs and CNNs performed alike and better than LogReg (Q3(CNN)=76.1-81.1% vs. Q3(LSTM)76.1-80.9%). As CNNs are computationally more efficient, we focused on those in the following.

*Comparing protein LMs:* Trained on UniRef100 (Table 1), ProtBert outperformed other models trained on the same corpus (Tables 3, 8). For ProtTXL and ProtBert we could analyze the influence of database size upon performance: 10-times larger

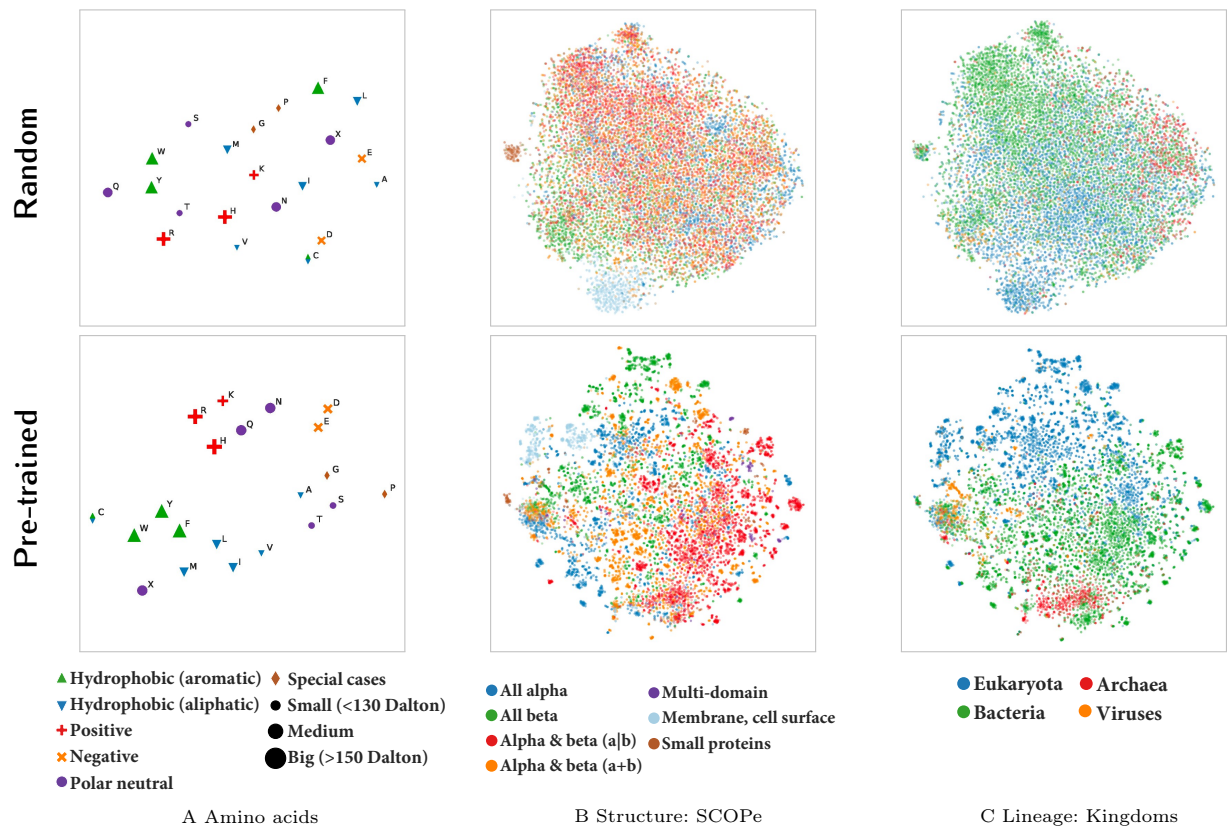

Fig. 4: Protein LMs learned constraints. t-SNE projections visualized information extracted by the unsupervised protein LMs (here best-performing ProtT5-U50; upper row: before training (Random), and lower row: after pre-training on BFD & UniRef50. (A) The left-most column highlights single amino acids by biophysical features. (B) The middle column annotates protein structural class (taken from SCOPe). (C) The right-most column distinguishes proteins according to the kingdom of life in which it is native. Although the random projections on top may suggest some adequacy of the clusters, the trained models shown on the lower row clearly stood out. Incidentally, the comparison of the two also highlighted the potential pitfalls of using t-SNE projections from many dimensions onto 2D: although random, the human might see some correct patterns even in the top row. Most impressive might be the fine-grained distinctions of biophysical features of amino acids (A), however, more surprising are the classifications of entire proteins according to structural class (B) and organism (C). For these, we created embeddings through global average pooling over the representations extracted from the last layer of ProtT5-U50 (average over protein length, i.e. per-protein embeddings; Fig. 1).

BFD (Table 1) helped ProtBert slightly ($\Delta$Q3: +1.1%) but made ProtTXL worse ($\Delta$Q3: -0.6%; Table 3 and SOM Tables 9, 8). The gain was larger when fine-tuning the two ProtT5 versions (XL and XXL) by first training on BFD and then refining on UniRef50. Consistently, all models fine-tuned on UniRef50 outperformed the versions trained only on BFD (Fig. 5, Table 3, SOM Table 8). Although these gains were consistently numerically higher, the statistical significance remained within the 68% confidence interval (maximal difference: 1.1% compared to one standard error of ±0.5%).

*Embeddings reach state-of-the-art (SOA):* All models (Prot-TXL, ProtBert, ProtAlbert, ProtXLNet, ProtElectra, ProtT5) and all databases (BFD, UniRef50/UniRef100) tested improved significantly over context-free feature extractors such as word2vec-based approaches (DeepProtVec in Fig. 5 and SOM Table 8). Both ProtTXL versions fell short compared to an existing ELMo/LSTM-based solution (DeepSeqVec [32]) while all other Transformer-models outperformed DeepSeqVec. Embeddings extracted from another large Transformer (ESMB-1b [72]), improved over all our non-ProtT5 models (Figs. 5 and SOM Table 8)). Most solutions using only embeddings as input were outperformed by the state-of-the-art method NetSurfP-2.0 [15] using evolutionary information (Fig. 5 and SOM Tables 9, 8).

However, ProtT5-XL-U50 reached nearly identical performance without ever using multiple sequence alignments (MSA). Analyzing the average Q3 per protein of both models for set NEW364 in more detail (SOM Fig. 12), revealed that 57% of the proteins were predicted with higher Q3 by ProtT5-XL-U50 (CASP12 was too small for such a differential analysis).

*LMs shine for small families:* The size of protein families follows the expected power-law/Zipf-distribution (few families have many members, most have fewer [80]). To simplify: families with fewer members carry less evolutionary information (EI) than those with more. One proxy for this is the number of effective sequences (Neff), i.e., the number of sequences in an MSA clustered at 62% PIDE [79], [81]. We analyzed the effect of Neff by comparing NetSurfP-2.0 (using MSAs/EI) to ProtT5-XL-U50 (not using MSAs) using four subsets of NEW364 using different Neff cutoffs, i.e., the subset of proteins without any hit (Neff=1, 12 proteins), less than 10 hits (Neff<=10, 49 proteins) and Neff>10 (314 proteins) (Fig. 6). Details on the MSA generation are given in SOM. ProtT5XL-U50 improved most over NetSurfP-2.0 for the smallest families (Neff=1).

*More samples better performance of protein LMs:* despite substantial differences in training corpus, hyperparameter choices and transformer model peculiarities, the LMs trained here exhibited a similar trend: correlating performance and the number of samples presented during the first step training of the protein LMs (*pre-training*). Toward this end, we computed the *number of samples* as the product of the *number of steps* and the *global batch size* (Fig. 7; Spearman's $\rho$=0.62). In particular, comparing the two largest models trained by us (ProtT5-XL and ProtT5-XXL) suggested that seeing more samples during pre-training might be more beneficial than increasing model size.

| Dataset | CASP12 | NEW364 |
|---|---|---|
| DeepProtVec | 62.9 | 64.7 |
| ProtTXL* | 71.5 | 72.8 |
| ProtTXL-BFD* | 71.7 | 72.2 |
| DeepSeqVec | 73.0 | 76.0 |
| ProtXLNet* | 73.7 | 77.3 |
| ProtElectra* | 73,9 | 78,1 |
| ProtAlbert* | 74.6 | 78.5 |
| ProtBert* | 75.0 | 80.1 |
| ProtBert-BFD* | 75.8 | 81.1 |
| ESM-1b | 76.9 | 82.6 |
| ProtT5-XXL-BFD* | 77.7 | 81.6 |
| ProtT5-XL-BFD* | 77.5 | 82.0 |
| ProtT5-XXL-U50* | 79.2 | 83.3 |
| ProtT5-XL-U50* | 81.4 | **84.8** |
| NetSurfP-2.0 | **82.0** | 84.3 |

TABLE 3: The three-state accuracy (Q3) for the per-residue/token-level secondary structure prediction (percentage of residues correctly predicted in either of 3 states: helix, strand, or other) for all protein LMs trained here (marked by star) along with other LMs, namely one word2vec-based approach (DeepProtVec), one LSTM (DeepSeqVec), one transformer (ESM-1b) and one of the current state-of-the-art methods (NetSurfP-2.0) that uses evolutionary information (EI)/multiple sequence alignments (MSAs). Values were compiled for two datasets: one because it is a standard in the field (CASP12, results for two other standard data sets - TS115 and CB513 - in Table 9), the other because it is larger and less redundant (dubbed NEW364 introduced here). Standard errors were computed using bootstrapping: CASP12=±1.6%, NEW364=±0.5%. Highest values in each column marked in bold-face.

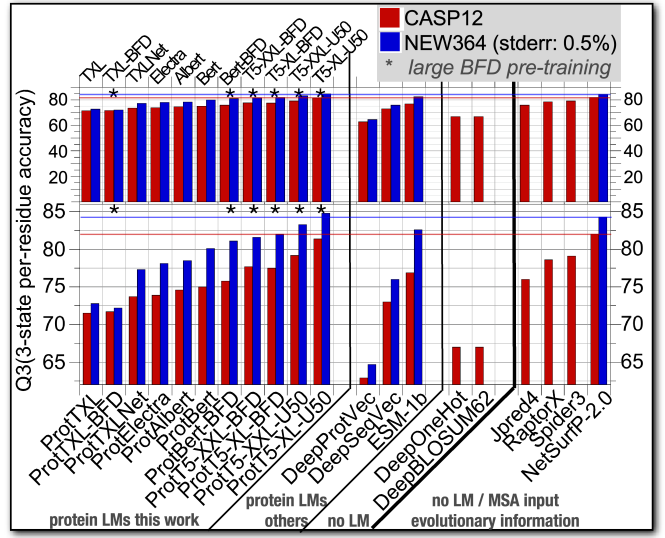

Fig. 5: Per-residue (token-level) performance for secondary structure prediction: CASP12 (red) and NEW364 (blue) constitute two test sets. Protein LMs trained here are shown in the left panel of the figure. Additions of BFD mark pre-training on the largest database BFD, U50 mark pre-training with BFD and refining with UniRef50. We included protein LMs described elsewhere (marked as: protein LMs others, namely ESM-1b [72], DeepProtVec and DeepSeqVec [32]). All embeddings were input to the same CNN architecture. Two approaches used amino acids instead of embeddings as input (marked as: no LMs: DeepOneHot [32] - one-hot encoding - and DeepBLOSUM62 [32] - input BLOSUM62 [74] substitution matrix), as well as, to the current state-of-the-art (SOA) method NetSurfP-2.0 [15], and Jpred4 [75], RaptorX [76], [77], Spider3 [78]. The rightmost four methods use MSA as input (marked as: MSA input evolutionary information). While only rotT5-XL-U50 reached the SOA without using MSAs, several protein LMs outperformed other methods using MSA. All protein LMs other than the context-free DeepProtVec improved significantly over methods using only amino acid information as input. One interpretation of the difference between the two data sets is that CASP12 provided a lower and NEW364 an upper limit. The top row shows the complete range from 0-100, while the lower row zooms into the range of differences relevant here.

### 3.2.2 Per-protein localization & membrane prediction

To investigate per-protein (sentence-level) aspects of protein function, we trained FNNs on sub-cellular localization (also referred to as *cellular compartment*) in ten classes and on the binary classification of membrane vs. non-membrane (also referred to as *globular*) proteins. Levels of ten-state (Q10 for localization) and two-state (Q2 for membrane/globular) measured performance. Toward this end, we derived per-residue embeddings from the last hidden layer of the protein LM and pooled over the length-dimension/entire protein (Fig. 1).

*Mean-pooling performed best:* Using ProtBERT-BFD embeddings we compared four different pooling strategies for collapsing per-residue (token-level) embeddings, the dimensions of which differ for proteins of different length, into representations of fixed length. These were min-, max-, and mean-pooling, as well as, the concatenation of those three (*concat*). The first two (min/max) performed almost fourteen percentage points worse for localization (Q10) and about three for membrane/other (Q2) compared to the others (mean/*concat*, Table 10). While mean-pooling and *concat* performed similarly for the classification task (membrane/other), mean-pooling outperformed *concat* for localization by about ten percentage points (Table 10). In the following, we used only mean-pooling to benchmark the per-protein/sentence-level predictions.

*Comparison of LMs:* the per-protein prediction of localization in 10 states largely confirmed the trend observed for per-residue secondary structure prediction (Q3/Q8) in several ways: All LMs introduced in this work (marked by * in Table 4) clearly outperformed the un-contextualized word2vec-based approaches (DeepProtVec; Fig. 8, Table 4). Except for ProtTXL and ProtXLNet, all transformers trained here outperformed the previous ELMo/LSTM-based solution (DeepSeqVec). Increasing the corpus for pre-training the protein LMs 10-fold appeared to have little effect (Prot* vs. Prot*-BFD in Fig. 8 and Table 4). In contrast, fine-tuning ProtT5 models already trained on BFD using UniRef50 improved (Prot*/Prot*-BFD vs. Prot*-U50 in Fig. 8 and Table 4). Although most embedding-based approaches were outperformed by the state-of-the-art method (*DeepLoc*) which uses multiple sequence alignments (MSAs) as input, both ProtT5 models trained on UniRef50 outperformed DeepLoc without using MSAs: Q10, Fig. 8 and Table 4.

*Similar for membrane/other:* Results for the classification into membrane/other (Q2; Table 4), largely confirmed those obtained for localization (Q10) and secondary structure (Q3/Q8):

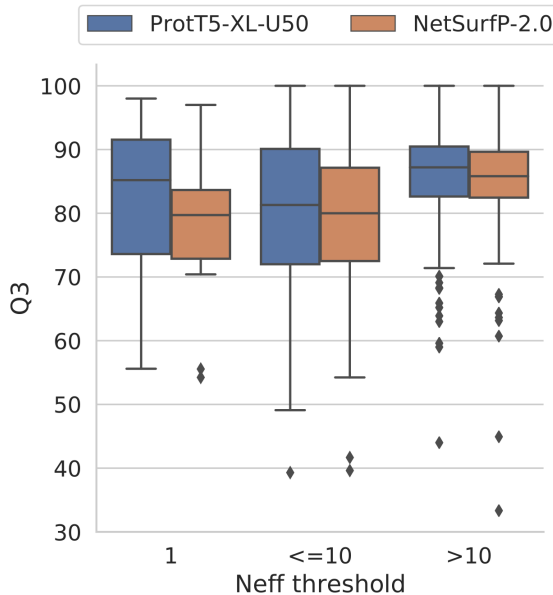

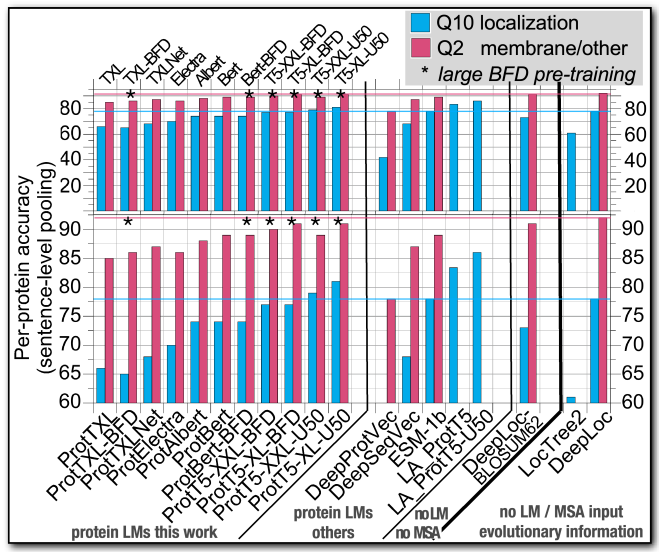

Fig. 6: Effect of MSA size. We used our new test set (NEW364) to analyze the effect of the size of an MSA upon secondary structure prediction (Q3) for the two top methods (both reaching Q3=84.3%): NetSurfP-2.0 (using MSA) and ProtT5-XL-U50 (not using MSA). As proxy for MSA size served Neff, the number of effective sequences [79] (clustered at 62% PIDE): leftmost bars: MSAs with Neff=1, middle: $Neff \leq 10$, right: Neff>10. As expected ProtT5-XL-U50 tended to reach higher levels than NetSurfP-2.0 for smaller families. Less expected was the almost on par performance for larger families.

Fig. 8: Per-protein (sentence-level) performance: The prediction of localization in 10 states (lower bars in cyan: Q10: percentage of proteins with 1 of 10 classes correctly predicted) and the classification of membrane/other (higher bars in magenta: Q2: percentage of proteins correctly classified in either of two classes). Embeddings were derived from protein LMs by mean-pooling, i.e. averaging over the length of the entire protein (Fig. 1). Abbreviations as in Table 5 except for one method using neither LMs nor MSA (no LM no MSA: DeepLoc-BLOSUM62 [16]), and two methods using MSAs (MSA input evolutionary information): the current state-of-the-art (SOA) method (performance marked by horizontal thin lines in magenta and cyan) DeepLoc [16], and LocTree2 [82]. Almost all LMs outperformed LocTree2 and a version of DeepLoc not using MSAs (DeepLoc-BLOSUM62). Only, ProtT5-XXL-U50 and ProtT5-XL-U50 outperformed the SOA. A recent method optimized localization prediction from embeddings (ProtT5) through a light-attention mechanism; it clearly outperformed the SOA without using MSAs (LA_ProtT5 & LA_ProtT5-U50 [83]). The top row shows the complete range from 0–100, while the lower row zooms into the range of differences relevant here.

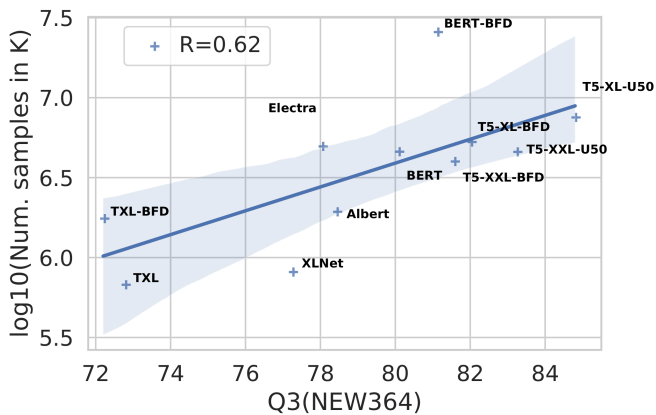

Fig. 7: Number of samples seen during pre-training correlates with performance - We compared 3-state secondary structure prediction performance (Q3) on NEW364 for all LMs trained here against the number of samples seen during pre-training (training steps times global batch-size in K). For simplicity, we dropped the prefix "Prot" from the models in this plot. Despite the peculiarities of each of the Transformers trained here, Spearman's $\rho$ of 0.62 indicates that there might be a common trend towards seeing more samples during pre-training.

(1) ProtT5 LMs fine-tuned on UniRef50 performed best without MSAs, (2) the 10-fold larger pre-training BFD had no noticeable effect, (3) our best protein LMs outperformed existing transformer LMs (ESM-1b) (Fig. 8). In contrast to localization and secondary structure, the fine-tuning appeared not to increase performance (Table 4) and both ProtT5 remained 1–2 percentage points below DeepLoc.

## 3.3 Fast and reliable predictions from embeddings

We compared the time needed to generate representations for EI-based prediction methods and protein language models by generating MSAs/embeddings for each protein in the human proteome (20,353) proteins with a median sequence length of 415 residues). We used the fastest method available, namely MMseqs2 [51], with parameters established by NetSurfP-2.0 to generate MSAs from two databases (UniRef90 with 113M and UniRef100 with 216M proteins; see SOM for more technical details). MMseqs2 was about 16 to 28-times slower than the fastest LMs (ProtElectra and ProtBert), and about 4 to 6-times slower than our best model (ProtT5-XL; Fig. 9). ProtT5-XL, required on average 0.12 seconds to generate embeddings for a human protein, completing the entire human proteome (all proteins in an organism) in only 40 minutes. We also investigated the cross-effect of sequence length and batch-size (SOM Table 11) on the inference speed of different protein LMs. When using a single Nvidia Quadro RTX 8000 with half precision on varying batch-sizes (1,16,32) as well as sequence lengths (128, 256, 512), ProtBert and ProtElectra provided the fastest inference with an average of 0.007 seconds per protein when using a batch size of 32, followed by ProtT5-XL and ProtAlbert (0.025s). The batch-size of most models could have been increased on the same hardware but was limited to allow a direct comparison between all models, due to large memory requirements for ProtT5-XXL. The script for running this benchmark is freely available[7].

| Dataset | Q10: Localization | Q2: Membrane/other |
|---|---|---|
| DeepProtVec | 42 | 78 |
| ProtTXL* | 66 | 85 |
| ProtTXL-BFD* | 65 | 86 |
| DeepSeqVec | 68 | 87 |
| ProtXLNet* | 68 | 87 |
| ProtElectra* | 70 | 86 |
| ProtAlbert* | 74 | 88 |
| ProtBert* | 74 | 89 |
| ProtBert-BFD* | 74 | 89 |
| ESM-1b | 78 | 89 |
| ProtT5-XXL-BFD* | 77 | 90 |
| ProtT5-XL-BFD* | 77 | 91 |
| ProtT5-XXL-U50* | 79 | 89 |
| ProtT5-XL-U50* | **81** | 91 |
| DeepLoc | 78 | **92** |

TABLE 4: Per-protein prediction of protein function: Given is the performance for two tasks that proxy the prediction of aspects of protein function, namely the prediction of subcellular localization (Localization) in ten states (Q10) and the classification of proteins into membrane-bound/other (Membrane/other) in two states (Q2). Values mark all protein LMs introduced here (marked by star) along with other LMs, namely one word2vec-based approach (Deep-ProtVec), one LSTM-based (DeepSeqVec), one transformer-based (ESM-1b) and the current state-of-the-art method (DeepLoc) that, unlike all other methods shown, used multiple sequence alignments (MSAs)/evolutionary information (EI) for the values shown. All values based on a standard, public data set [16]. Highest values in each column marked in bold-face.

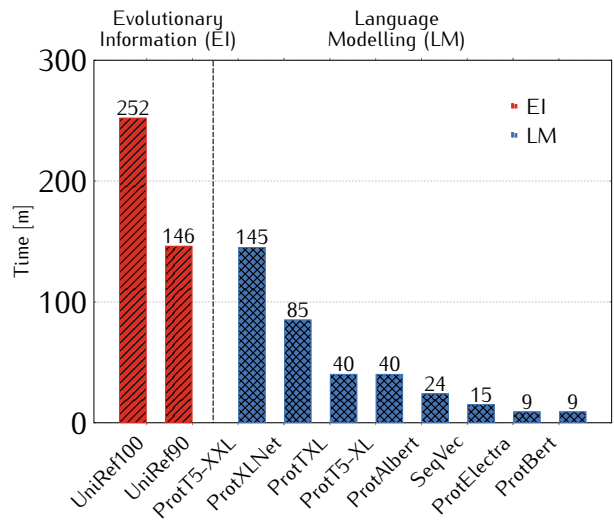

Fig. 9: Inference Speed Comparison: The time required to generate protein representations for the human proteome (20.353 proteins) is compared using either our protein LMs or mmseqs2 (protein sequence search tool [51] used to generate evolutionary information; NetSurfP-2.0 [15] parameters are used). Here, we used mmseqs2 (red bar) to search each protein in the human proteome against two large protein database (UniRef90 and UniRef100 with 113M and 216M proteins, respectively). Only embedding or search time is reported, i.e. no pre-processing or pre-training was measured. mmseqs2 was run on an Intel Skylake Gold 6248 processor with 40 threads, SSD and 377GB main memory, while protein LMs were run on a single Nvidia Quadro RTX 8000 with 48GB memory using half precision and dynamic batch size depending on sequence length (blue bar).

## 4   DISCUSSION

### 4.1   *Substantial computing resources needed to cope*

HPC Supercomputers such as Summit [1] and Google's cloud TPU Pod [2], combined with optimized libraries such as IBM DDL [7] and Horovod [6] set the stage for training LMs with billions of free parameters on corpora with terabytes of data in hours or days. Increasing model size improves performance for some NLP applications [12], although the massive data challenges the communication between thousands of nodes and divergence between large batches during training. Here, we presented some solutions to overcome these challenges for training protein LMs by fully utilizing 20% of Summit for TransformerXL [59], as well as, by using one TPU Pod V3-512 for Bert [57], Electra [56], Albert [58] and XLNet [11], and a mix of TPU Pod V3-256, V3-512, and V3-1024 for ProtT5-XL and ProtT5-XXL [55]. This translated into the parallel use of 5616 GPUs on Summit or 256/512/1024 TPU cores on a TPU Pod, while avoiding training divergence with specialized optimizers such as LAMB [61] up to a global batch size of 44K samples (here: protein sequences).

### 4.2   Training protein LMs longer most important

*Better protein LM means higher performance when using it as input.* For most of our work, we proxied the degree to which the pre-trained protein language models (LMs) extracted information through the performance of the second step supervised tasks. Consequently, we considered that the *protein LM improved (or was better)* when the supervised task using this LM as input reached higher performance.

*BFD: Largest database for pre-training:* We trained our protein LMs on the largest protein database ever used for this purpose, namely BFD [42], more than an order of magnitude larger than UniProt [23], the standard in the field. Although bigger did not equate better in terms of prediction performance on supervised tasks, some of the protein LMs appeared to improve through pre-training on more data (UniRef100 vs BFD, Table 1). Nevertheless, the top performance increase appeared somehow limited given the 10-fold larger data set (e.g. ΔQ3(ProtBert-BFD/UniProt)=1.3%). Instead, the pre-training fine-tuning protocol by which we first trained on the larger but more noisy (more mistakes in protein sequences) and redundant BFD and then continued pre-training using the smaller, less redundant UniRef50 improved performance significantly for both ProtT5 versions (ΔQ3(ProtT5-XL-BFD/U50)=2.8% and ΔQ3(ProtT5-XXL-BFD/U50)=1.4%). Possibly, the refined models were less biased toward large protein families over-represented in redundant databases. The improvement through refined pre-training for ProtT5-XL (3 parameters) exceeded that for ProtT5-XXL (11B parameters), presumably, because it saw more samples when continuing pre-training for a similar amount of time (limited by resources).

This highlighted a remarkable trend common across an immense diversity of protein LMs and corpus: the performance of supervised downstream tasks using the embeddings from pre-trained protein LMs as input increased with the number of samples presented during the LM pre-training (Fig. 7; Spearman's

7. https://github.com/agemagician/ProtTrans/tree/master/Benchmark

$\rho$=0.62). We could not observe a similarly consistent trend for model size. However, this might be attributed to some trade-off between both: training for more steps might also require a model with sufficient capacity to absorb the information of the corpus. For instance, while ProtBERT-BFD (420M parameters) saw around 27B proteins during pre-training, it fell short compared to ProtT5-XL-BFD (3B parameters) which saw only around 5B proteins (Figs. 5, Table 3, and SOM Table 8). This finding appeared to confirm results from NLP suggesting that larger models absorb information faster and need less training time to achieve similar performance [84]. However, the comparison between, e.g., ProtT5-XL and ProtT5-XXL suggested a possible cap to this trend as larger models saw fewer samples in the same amount of computing power. The clear correlation between performance and samples seen during pre-training combined with the need for sufficient model size spotlighted the crucial role of substantial computing resources (HPC, TPUs, and GPUs): big data needs large models need loads of computational resources.

### 4.3 Protein LMs learned global constraints

Some rudimentary information about how proteins are formed, shaped, and function has been learned by the protein LMs during pre-training because all models (ProtT5, ProtBert, ProtElectra, ProtAlbert, ProtTXL, ProtXLNet) extracted valuable information as revealed by visualizing embeddings without further supervised training on labeled data. The comparison to a random LMs highlighted two important aspect. Firstly, how easy it is to mis-interpret patterns when projecting from high-dimensional spaces upon 2D: although the randomly initialized protein LMs contained no information, some annotations might have suggested the opposite (top row in Fig. 4 learned NO information). Secondly, the protein LMs did extract important global constraints relevant for protein structure and function (lower row in Fig. 4). This span from the most local (individual token level) biophysical features of the amino acid building blocks (e.g. hydrophobicity, charge, and size, Fig. 4A), over global classifications of proteins according to structural classes (Fig. 4B), to the macroscopic level of the domains of life (Fig. 4C). Global structural properties (e.g. overall secondary structure content, Fig. 4B) and global biochemical properties (e.g. membrane-boundness, SOM Fig. 14B) appeared most distinctive. In contrast, local features relying on short motifs were less separated (EC-numbers: Fig. 14F, localization: Fig. 14C) but still clustered, e.g., for secreted (extracellular) proteins or hydrolases.

On a more fine-grained level, the visual analysis of the attention mechanism at the core of each of the transformer models trained here, confirmed that the protein LMs even picked up more subtle signals of short functional motifs. Specifically, one of the attention heads of ProtAlbert zoomed mostly into the four residues most important for the coordination of zinc-binding (SOM Fig. 11). Although limited in scope [71], such an analysis provides some explanation about the inner workings of the Transformer models without needing large sets of experimental annotations (labels). On top, the resulting *interpretations of the AI/ML* might be less biased than experimental annotations. For instance, databases with annotations of protein function such as Swiss-Prot [85] and of protein structure such as PDB [50] are extremely biased by what today's experimental techniques can handle [80], [86], [87].

### 4.4 Protein LMs top without MSAs

The t-SNE and UMAP analyses suggested that the protein LMs had extracted some level of *understanding of the language of life*. However, prediction is the acid test for understanding. To pass this test, we extracted the embeddings learned by the protein LMs directly as input to predict aspects of protein structure (per-residue/token-level prediction of secondary structure) and protein function (per-protein/sentence-level prediction of localization and membrane/other). Overall, the results obtained from the second step of using embeddings from LMs as input to supervised models confirmed [32] that evolutionary information (EI, i.e. methods using multiple sequence alignments MSAs) scientifically and statistically significantly outperformed most LMs not using such information except for ProtT5-XL (on all per-residue and per-protein tasks, Figs. 5, 8 and SOM Tables 9, 8). ProtT5-XL eliminated this gap from embeddings-only input: on some tasks/data sets, it outperformed the current state-of-the-art MSA-based method, on others it remained slightly behind. Newer protein LMs using *context* improved over both previous LM-based approaches [32] (8-9 percentage points in Q3), other transformers [72] (2-4 percentage points in Q3), and over non-contextualized word2vec-type approaches [88], [89], [90] (18-22 percentage points in Q3). The performance ranges for using two different data sets (CASP12 and NEW364) highlight a different problem. While it is clear that we need to redundancy-reduce evaluations sets with respect to themselves and all data used for development, it is less clear how to exactly do this. In focusing on CASP12 and NEW364, we approached two different assumption. CASP12 is best described as measuring how well predictions will be for proteins with very different structures. A comprehensive rigorous realization of data sets following this perspective has recently been published [91]. NEW364, on the other hand, builds on the assumption that the maximal redundancy is defined by *sequence similar to protein in the PDB*. In this sense, we interpreted results for CASP12 as a lower and those for NEW364 as an upper limit. Either way, the most important task is to constantly create up-to-date sets with enough non-redundant proteins never used for development by any of the methods assessed.

*Protein LMs so powerful that even simple baseline are effective:* While pre-training is computationally demanding, training supervised models using embeddings as input requires much fewer resources. For instance, the logistic regression trained on top of ProtBERT-BFD was already competitive with substantially more complex CNNs or LSTMs in predicting secondary structure (SOM Table 7). In another example, a parameter-free nearest neighbor lookup using distances from protein LM embeddings sufficed to outperform homology based inference for predicting protein function [92]. This suggested that protein LMs are particularly suitable when the experimental annotations (labels) available are very limited and hinder training of large supervised networks. In fact, none of the supervised solutions presented here that reached the SOA came anywhere near in complexity (number of free parameters) to that of the EI-based methods they reached. Here, we focused on the development of protein LMs and used performance on supervised tasks primarily as a proof of principle without optimizing particular supervised solutions. Others have already begun beating EI-based methods not using protein LMs by custom-designing such solutions [83] possibly even through end-to-end systems [30], [38]. Combinations of evolutionary information and embeddings might bring the most accurate methods. However, such a merger would sacrifice the advantage of protein LMs relying only on single protein sequences which is crucial for large scale analysis as well as certain tasks like single amino acid variant (SAV) effect prediction.

*Bi-directionality crucial for protein LMs:* In NLP uni-directional (auto-regressive) and bi-directional (auto-encoding) models perform *on par* [12], [93]. In contrast, the bi-directional context appeared crucial to model aspects of the language of life. While auto-encoding models such as Albert [58] utilize context to both sides during loss calculation, auto-regressive models such as TransformerXL [59] consider only context to one side. Performance increased substantially from uni-directional ProtTXL to bi-directional ProtXLNet (Fig. 5, Table 3, and SOM Table 8). This might be compensated for by first pre-training on sequences and their reverse and then concatenating the output of uni-directional LMs applied on both directions. While this does not allow the LM to use bi-directional context during training, it allows supervised networks to combine context derived independently from both sides. For instance, ELMo [8] concatenates the embeddings derived from a forward and a backward LSTM. The protein LM version of ELMo (SeqVec) outperformed the uni-directional ProtTXL but not the bi-directional ProtXLNet. The difference in model size (SeqVec=93M vs. ProtXLNet=409M) and in pre-training data (SeqVec=30M vs. ProtAlbert=224M) might explain some of this effect. Nevertheless, pure uni-directionality as used in TransformerXL appeared detrimental for modeling protein sequences.

## 4.5   Have protein LMs reached a ceiling?

Applying techniques from NLP to proteins opens new opportunities to extract information from proteins in a self-supervised, data-driven way. New protein representations may complement existing solutions, most successful when combining evolutionary information[8] and machine learning [21], [22], [40], [94]. Here we showed for the first time that embeddings from protein LMs input to relatively simple supervised learning models can reach similar levels of performance without using EI and without optimizing the supervised training pipeline much. However, the gain in inference speed for protein LMs compared to traditional models using evolutionary information is so significant that large-scale predictions become, for the first time since 30 years, feasible on commodity hardware. For instance, the best-performing model ProtT5-XL-U50 can run on a Nvidia TitanV with 12GB vRAM (see Methods for details). Nevertheless, given the experiments described here and in previous work [32], [33], [34], [35], [36], [37], [39], we might expect an upper limit for what protein LMs can learn when using masked language modeling (or auto-regressive pre-training) exclusively. Although this work explicitly addressed the possibility of reaching such a limit, we could not conclusively provide an answer. We could establish three findings. (1) Less noisy and less redundant corpora (e.g. UniRef50) improved over larger but more noisy and redundant corpora (e.g. BFD). (2) In our perspective of limited resources, it was most important to use the resources for long-enough training because the number of samples seen during pre-training correlated with the prediction performance of downstream tasks. Ultimately, this seemed to originate from a trade-off between sufficient model size and sample throughput. (3) The bi-directional outperformed the uni-directional models tested. However, given the advances of protein LMs over the course of the reviewing of this work, we have seen no evidence for having reached a limit for protein LMs, yet.

*Many open questions:* Answers to the following questions might advance the status-quo. **(1)** Would the addition of auxiliary tasks such as next-sentence or sentence-order prediction offered by BERT or Albert suit protein sequences? A suggestion might be the usage of structure information [95] or evolutionary relationship [35], [96]. **(2)** Might the efficiency of transformers protein LM training improve through sparse transformers [97] or attention optimized with locality-sensitive hashing (LSH) [98] as introduced recently by the Reformer model [99] or more recent work of linear transformers [100]? **(3)** Which data set prepossessing, reduction and training batch sampling should optimally used for better results? **(4)** How much will it improve to tailor the supervised training pipeline to particular tasks? We treated secondary structure or localization prediction more as proxies to showcase the success of protein LMs than as an independent end. **(5)** Will the combination of EI and AI [96] bring the best protein predictions of the future, or will the advantages of single-protein predictions (speed, precision) win out? In fact, single-protein predictions also have the advantage of being more precise in that they do not provide *some implicit average over a protein family.*

Overall, our results established that the combination of HPC solutions for training protein LMs and subsequent training of supervised prediction methods scaled up to the largest data sets ever used in the field. Only the combination of these different domains allowed us to demonstrate that protein LMs can reach up to the same performance of the state-of-the-art of methods combining EI and AI without ever exploiting multiple sequence alignments.

## 5   CONCLUSION

Here, we introduced many novel protein language models (LMs) and proved that embeddings extracted from the last LM layers captured constraints relevant for protein structure and function. Although neither the usage of the largest ever database for a protein LMs (BFD), nor that of very large models generated the most informative embeddings, pre-training sufficiently long on considerable diversity made a difference, and more recent LMs performed best. Using embeddings as exclusive input to relatively small-size CNN/FNN models without much optimization yielded methods that appeared competitive in predicting secondary structure, localization and in classifying proteins into membrane/other. In fact, for the first time, new small-size supervised solutions based on LMs embedding input reached levels of performance challenging the state-of-the-art (SOA) methods based on multiple sequence alignment (MSA) input. In contrast, the models presented here never used MSAs. This could save immense expenses when routinely applying embedding-based protein predictions to large data sets, but it also opens a path toward protein-specific rather than family-averaged predictions. Ultimately, joining the strengths of three different, yet complementary, fields (HPC, NLP and computational biology) affected the advance. Self-supervised pre-training combined with transfer-learning tapped into the gold-mines of unlabeled data opening the door for completely novel perspectives (and solutions) on existing problems.

---

8. Throughout this work, we used evolutionary information (EI) as synonymous for *using multiple sequence alignments (MSAs).* Whether protein LMs do not implicitly extract EI will have to be proven in separate publications.

## 6 AVAILABILITY

We made all protein LMs trained here publicly available at our ProtTrans repository "https://github.com/agemagician/ProtTrans/". This repository also holds jupyter python notebooks with various tutorials, e.g., on how to extract embeddings or visualize attention using freely available online resources (Google Colab).

### Acknowledgments

The authors thank primarily Tim Karl (TUM) and Jian Kong (TUM) for invaluable help with hard- and software; Inga Weise and Aline Schmidt (both TUM) for support with many other aspects of this work; Florian Matthes (TUM) for his generous support and encouragement. Thanks for crucial support and feedback from NVIDIA, in particular to Ulrich Michaelis, Ada Sedova, Geetika Gupta, Axel Koehler, Frederic Pariente, Jonathan Lefman, and Thomas Bradley. Thanks to many at ORNL without whom no aspect of this work could have been realized; particular thanks to John Gounley, Hong-Jun Yoon, Georgia Tourassi, Bill, Brian, Junqi, Graham and Verónica (ORNL Summit). Furthermore, special thanks to Jack Wells (ORNL) for opening the door to kicking off this project. From IBM, we thank Nicolas Castet and Bryant Nelson for their help to fix issues and enhance the performance of IBM PowerAI. From Google, we are deeply grateful to Jamie Kinney, Alex Schroeder, Nicole DeSantis, Andrew Stein, Vishal Mishra, Eleazar Ortiz, Nora Limbourg, Cristian Mezzanotte and all TFRC Team for helping to setup a project on Google Cloud and solving Google cloud issues. No ProtTrans model were easily publicly available without support from the Hugging Face team; including Patrick von Platen, Julien Chaumond, and Clement Delangue. Special thanks to Konstantin Schütze for helping with grant writing and providing early results for the structure prediction task. Furthermore, thanks to both Adam Roberts and Colin Raffel for help with the T5 model. We are grateful to the editor and the anonymous reviewers for essential criticism, especially, for suggesting to compare t-SNEs to randomly initialized models.

This work was supported by a grant from Software Campus 2.0 (TUM) through the German Ministry for Research and Education (BMBF), a grant from the Alexander von Humboldt foundation through the German Ministry for Research and Education (BMBF), and by a grant from the Deutsche Forschungsgemeinschaft (DFG–GZ: RO1320/4–1). We gratefully acknowledge the support of NVIDIA with the donation of 2 Titan GPUs used for the development phase. We also thank the Leibniz Rechenzentrum (LRZ) for providing access to DGX-1(V100) for the testing phase. Martin Steinegger acknowledges support from the National Research Foundation of Korea grant [2019R1A6A1A10073437, NRF-2020M3A9G7103933]; New Faculty Startup Fund and the Creative-Pioneering Researchers Program through Seoul National University.

Last not least, this research used resources of the Oak Ridge National Laboratory (ORNL) Leadership Computing Facility, which is a DOE Office of Science User Facility supported under Contract DE-AC05-00OR22725, and resources of TPU pods under TensorFlow Research Cloud grant. Furthermore, the Rostlab gladly acknowledges support from Google Cloud and Google Cloud Research Credits program to fund this project under Covid19 HPC Consortium grant.

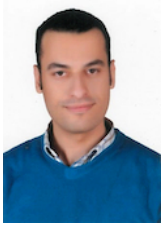

**Ahmed Elnaggar** is a PhD candidate at the Technical University of Munich. His main focus of research is self-supervised learning on various modalities (Text, Protein, Source code, Images and speech) using high performance computing.

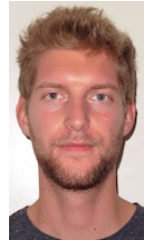

**Michael Heinzinger** is a PhD candidate in the Rostlab at TUM in Munich/Garching. His recent focuses on learning, evaluating and understanding representations for protein sequences from unlabeled data with the goal to empower peers with the computational tools necessary to unravel more fundamental biological truths.

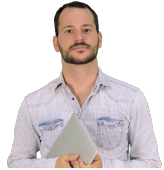

**Christian Dallago** performs research at the interface of Biology, Machine Learning and Software Engineering with the goal of improving human health through intelligent machines.

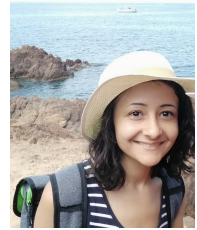

**Ghalia Rehawi** is a PhD candidate at Helmholtz Zentrum München. She completed her Master of Science (Msc), in the field of informatics, from the Technical University of Munich. She is interested in the application of machine and deep learning techniques in genome and transcriptome analysis.

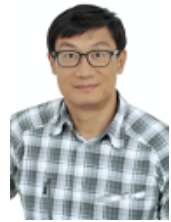

**Yu Wang** studied AI at Katholieke Universiteit Leuven in Belgium. He later moved to Munich, Germany to join MIPS, Helmholtz Zentrum München, where he got his Ph.D. in genomics and bioinformatics from Technical University Munich in 2011. He is currently CTO of Med AI Technology (Wu Xi) Ltd., working on transforming healthcare with AI in China.

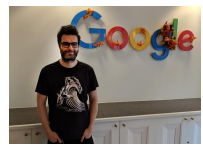

**Llion Jones** is a senior software engineer at Google and has been at Google for over 9 years. Started as a YouTube engineer before moving into machine learning. Was on the original team of researchers who developed the now popular Transformer model where he worked on the initial code base and on the attention visualizations.

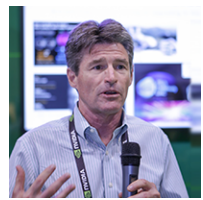

**Tom Gibbs** manages Developer Relations for NVIDIA's Supercomputing Business Unit. His primary focus areas are AI for Science, the Convergence of Simulation and Experiment, Quantum Computing and Classical Simulation of High Energy Physics. He has over 40 years of experience in large scale simulation and modeling with an emphasis on grand challenge science problems.

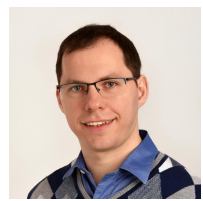

**Tamas Feher** is an AI developer technology engineer at NVIDIA. His work is focused on accelerating deep learning and machine learning workloads on GPUs. Tamas holds a PhD from the University of Greifswald.

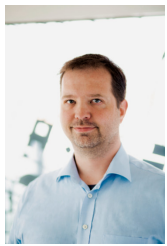

**Christoph Angerer** is a senior manager within the Autonomous Driving team at NVIDIA. Christoph's team is concerned with designing, implementing, and optimizing AI-based solutions for advanced learning and automation. Christoph holds a PhD from the ETH Zurich, Switzerland.

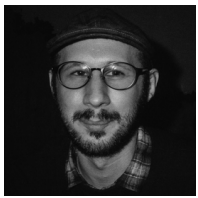

**Martin Steinegger** is an Assistant Professor in the biology department at the Seoul National University. His group develops novel computational methods that combine big data algorithms and machine learning to gain insights into unexplored microbial communities.

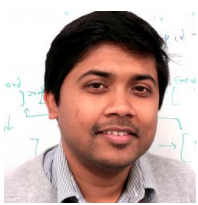

**Debsindhu Bhowmik** is a Computational Scientist in the Computational Sciences & Engineering Division and Health Data Sciences Institute at Oak Ridge National Laboratory. His current focus is in understanding complex biological and genetic phenomena and studying disordered systems by implementing new generation large scale simulation blended with Deep learning and scattering techniques.

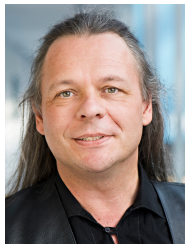

**Burkhard Rost** chairs *Comp Biol & Bioinformatics* at TUM Munich. He rooted the leap through combining evolutionary information and machine learning and the launch of *PredictProtein* as first Internet prediction server. Over 30 years, the *Rostlab* contributed influential methods for protein prediction, headed the International Society for Computational Biology (ISCB) and has been dedicated to teaching and raising diversity and gender balance.

# SUPPLEMENTARY ONLINE MATERIAL (SOM)

## -1.1 Datasets

**Language model corpora.** Here, we show more details on the differences between the different corpora used for protein LM pre-training. Towards this end, we compare the number of sequences, residues as well as the amino acid distribution in UniRef50, UniRef100 and BFD. We also compare the storage size required after converting the protein sequences in each of the databases to tensorflow records (SOM Fig. 10).

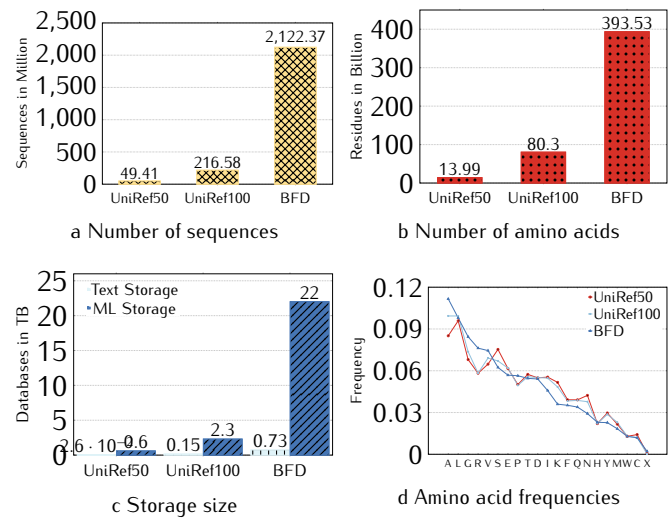

a Number of sequences

b Number of amino acids

c Storage size

d Amino acid frequencies

Fig. 10: Large Scale Dataset Training: here we compare the three datasets that were used in this study for language modelling (UniRef50, UniRef100, BFD). a) shows the number of sequences in each dataset in millions. (b) shows the number of residues/tokens in each dataset in billions. (c) shows size of each dataset raw text files as well as after converting to tensors in terabytes. (d) shows the frequency of each amino-acid/token in the each dataset

**Secondary structure data sets.** We also give a detailed overview of the three– and eight–state secondary structure class distribution for the NetSurfP–2.0 [15] training set (Train), the three test sets used in NetSurfP–2.0 (CASP12 [49], TS115 [48], CB513 [47]) as well as our new test set (NEW364) in SOM Tables 6, 5.

| Dataset | H | E | - | S | T | G | B | I |
|---------|------|------|--------|------|-------|-------|-------|-------|
| CASP12 | 1989 | 1416 | 1400 | 668 | 633 | 215 | 62 | 37 |
| TS115 | 10434 | 5085 | 5395 | 2210 | 2875 | 1033 | 295 | 174 |
| CB513 | 25559 | 12585 | 17713 | 8211 | 9711 | 3074 | 1105 | 469 |
| NEW364 | 26182 | 16563 | 14911 | 6233 | 7923 | 2732 | 797 | 15 |
| Train | 888175 | 595370 | 504272 | 204830 | 282561 | 99799 | 26420 | 14178 |

TABLE 5: Class distribution 8-state secondary structure - A detailed overview of the class distribution for the secondary structure datasets in 8-states is given. We compare the original NetSurfP-2.0 training set (Train), the corresponding validation datasets (CASP12, TS115, CB513) and our new test set (NEW364).

## -1.2 Step 1: Protein LMs unsupervised

The protein LMs trained here can also be used without any supervised training either by performing a qualitative analysis of the embedding space by projecting the high-dimensional representations for a set of proteins to 2D using e.g. t-SNE. Towards this end, proteins from SCOPe (v2.07 clustered at 40% PIDE) were used as proxy for protein structure (dubbed *SCOPe* in SOM Fig14–19), the three kingdoms of life (dubbed *Lineage*) and Virus as well as protein function (dubbed *Function*). For the functional classification we used only the subset of experimentally annotated EC (Enzyme Commission [5]) numbers removing all proteins that were not mappable. Protein function was also analyzed from an orthogonal angle using a different, 30%PIDE non-redundant, protein set [16] that allowed to analyze whether or not the embeddings captured aspects of the *cellular compartment; dubbed Localization in SOM Fig14–19* and membrane-association(dubbed *Membrane vs Soluble*). Additionally, we visualized the attention mechanism for a single protein to highlight the different ways and scales that Transformers offer to analyze proteins.

| Dataset | Proteins | Helix | Strand | Irregular |
|---------|----------|--------|---------|-----------|
| CASP12 | 21 | 1478 | 2241 | 2701 |
| TS115 | 115 | 5380 | 11641 | 10480 |
| CB513 | 511 | 18690 | 29102 | 35635 |
| NEW364 | 363 | 17360 | 28929 | 29067 |
| Train | 10796 | 585790 | 1002152 | 996673 |

TABLE 6: Class distribution 3-state secondary structure - An overview of the class distribution for 3-state secondary structure data sets used in this work is given. We compare the original NetSurfP-2.0 training set (Train), the corresponding validation datasets (CASP12, TS115, CB513) and our new test set (NEW364).

**Visual embedding space analysis** - Using t-SNE projections, the information content stored within the novel embeddings was qualitatively assessed on various levels, ranging from bio-physical and bio-chemical properties of single amino acids over different aspects of protein function (E.C. numbers, subcellular localization and membrane-boundness) to the level of kingdoms of life, i.e. Eukaryota, Bacteria and Archaea (for completeness here also including Viruses). We have visualized those different protein modalities for a subset of our language models, i.e. for ProtT5-U50 (SOM Fig. 14D), ProtBERT-BFD (SOM Fig. 15D), ProtBERT (SOM Fig. 16D), ProtAlbert (SOM Fig. 17D), ProtXLNet (SOM Fig. 18D) and ProtTXL (SOM Fig. 19D). When interpreting visual entropy in those figures as a proxy for the information learnt by the models, we can observe a similar trend than for the supervised comparison, i.e. ProtT5 seems to learn on all levels a better clustering than ProtBERT.

**Attention mechanism visualization** - In line with the idea of explainable AI which tries to move away from the black-box stereotype of neural networks towards understanding why a model made a certain prediction, the analysis of the attention mechanism [70] that is at the core of each Transformer model [10] allows, to a certain extent [71], to draw first conclusions about the inner workings and the resulting predictions of Transformers. Applied to protein sequences, it was shown that the attention mechanism of Transformers can be used to predict contacts between residue pairs that are close in 3D space but far apart in sequence space [72]. Here, we visualize the attention weights [73] of one of the Transformers trained here (ProtAlbert) to analyze the structural motif of a zinc-binding domain (SOM Fig. 11). This structural motif is crucial for DNA and RNA binding across a multitude of organisms. In order to coordinate the overall fold of zinc-fingers, the binding of four specific residues, usually two Cysteines and two Histidines, to a zinc-ion is crucial. Due to their importance for the correct functioning of this protein, these residues are well conserved across most organisms with zinc-finger domains. The visual analysis of the attention triggered by the first 33 residues of an exemplary zinc-finger (PDB: 1A1L [101]) confirms that one of the attention heads in the fifth layer of ProtAlbert mostly attends to the four residues involved in coordinating the zinc-binding which indicates that the model could have learnt to pick up the signal of this, relatively frequently occurring, structural motif. Such an analysis could allow for a cheap and fast analysis of single proteins without a) needing large labeled datasets for supervised training and b) being less influenced by the experimental bias in today's labeled databases which focus mostly on model organisms with applications to biotechnology.

## -1.3 Supervised Learning

**Supervised architecture comparison.** In this section we compared a) different choices for the supervised network that we use to evaluate the secondary structure prediction performance of the language models trained here (Table 7), b) performance of our ProtTrans models on established datasets to simplify comparability to existing prediction methods (SOM Tables 9, 8) and c) different choices for pooling embeddings of variable length protein sequence embeddings to a fixed-size representation that can be used for classifying whole protein sequences (Table 10) and

**Supervised architecture comparison.** Different architecture choices for the supervised network used to make predictions on the token-level are possible , i.e., various networks can be used to predict secondary structure for each amino acid in a protein. Towards this end, we used embeddings derived from ProtBERT-BFD to evaluate one linear classifier (logistic regression) and three non-linear classifiers (FNN, CNN and LSTM) on secondary structure prediction performance in three- and eight-states using the two hardest test sets, i.e., CASP12 and our new test set (SOM Table 7). This analysis revealed that even a logistic regression achieves competitive performance, indicating that secondary structure information is readily available from ProtBERT embeddings. Adding non-linearity without considering neighboring token embeddings (FNN) improves prediction performance and allowing the supervised network to harness local neighboring information (CNN) improves results further. However, allowing the supervised network to learn more long-range information (LSTM) only lead to an insignificant improvement for one out of four benchmarks (Q8(CASP12)) while adding more computational complexity due to the sequential nature of LSTMs. This shows that a) Transformer models already capture most of the long-range information removing the necessity to apply LSTMs, b) applying CNNs instead of FNNs improves performance slightly, potentially due to the inductive local bias of CNNs that fits well to the local nature of secondary structure. Therefore, we used a CNN architecture to evaluate the secondary structure prediction performance of all language models trained here.

**Per-residue (token level) prediction of secondary structure.** All models were evaluated using standard measures for performance (Q3/Q8: three/eight-state per-residue accuracy, i.e. percentage of residues predicted correctly in either of the 3/8 states) on standard datasets (CASP12, TS115, CB513) and a novel, highly non-redundant test set (NEW364). While TS115 and CB513 might overestimate performance because they allowed for more redundancy, we added Q3 on TS115 and CB513 to SOM Table 9 to ease comparability to existing approaches. As Q8 largely confirmed the trend observed for Q3, we added Q8 for all sets to SOM Table 8.

| Model | Q3(CASP12) | Q3(NEW364) | Q8(CASP12) | Q8(NEW364) |
|---|---|---|---|---|
| CNN | **76.1** | **81.1** | 65.2 | **70.3** |
| FNN | 75.3 | 80.0 | 63.6 | 68.9 |
| LSTM | **76.1** | 80.9 | **65.6** | 70.0 |
| LogReg | 74.3 | 79.3 | 63.4 | 68.1 |

TABLE 7: Architecture choice - We used one of our language models (ProtBERT-BFD) to compare different choices for the supervised network that we train to make predictions on the level of single amino acids, i.e, we compared a CNN, FNN, LSTM, and a logistic regression on 3- and 8-state secondary structure prediction on two different test sets (CASP12 and our new test set). Overall, the CNN provides the best performance while being computationally more efficient than the LSTM which reaches a similar performance. Despite its lower expressive power, even the logistic regression achieves competitive performance highlighting that the language models introduced here already learnt aspects of secondary structure during pre-training.

| Dataset | CASP12 | TS115 | CB513 | NEW364 |
|---|---|---|---|---|
| DeepProtVec | 49.7 | 54.4 | 48.9 | 53.3 |
| ProtTXL* | 58.5 | 63.3 | 58.9 | 61.0 |
| ProtTXL-BFD* | 58.6 | 63.3 | 58.8 | 60.9 |
| DeepSeqVec | 61.0 | 67.2 | 62.7 | 64.8 |
| ProtXLNet* | 61.6 | 68.6 | 63.1 | 65.6 |
| ProtElectra* | 60.9 | 69.1 | 64.7 | 66.9 |
| ProtAlbert* | 62.1 | 69.9 | 64.9 | 66.9 |
| ProtBert* | 63.3 | 71.5 | 66.6 | 68.9 |
| ProtBert-BFD* | 65.1 | 73.3 | 69.6 | 70.4 |
| ESM-1b | 66.0 | 73.4 | 70.2 | 71.3 |
| ProtT5-XXL-BFD* | 66.0 | 73.4 | 69.8 | 70.5 |
| ProtT5-XL-BFD* | 66.4 | 74.2 | 71.0 | 71.2 |
| ProtT5-XXL-U50* | 68.1 | 75.1 | 71.6 | 72.5 |
| NetSurfP-2.0 | 70.3 | 75.0 | 72.3 | 73.9 |
| ProtT5-XL-U50* | **70.5** | **77.1** | **74.5** | **74.5** |

TABLE 8: 8-state secondary structure prediction performance (Q8) - A detailed overview of the accuracy for correctly predicting secondary structure in 8-states is given here. We compare the performance of all language models trained here (marked with star), one word2vec-based approach (DeepProtVec), one LSTM-based (DeepSeqVec), one Transformer-based (ESM-1b) and one of the current state-of-the-art approaches that utilizes evolutionary information (NetSurfP-2.0) on three existing validation datasets (CASP12, TS115, CB513) and our new test set (NEW364). Standard errors were computed using bootstrapping: CASP12=±1.9, TS115=±1.0,CB513=±0.6,NEW364=±0.7.

Additionally, we added a detailed comparison of the 3-state secondary structure prediction performance (Q3) on NEW364 between NetSurfP-2.0 and our best performing protein LM (ProtT5-XL-U50; SOM Fig. 12). Each dot in the scatter plot reflects the Q3 achieved by NetSurfP2.0 and ProtT5-XL-U50 for the same protein. While using only single protein sequences, ProtT5-XL-U50 outperforms NetSurfP-2.0 for 57% (208 out of 364) proteins.

**MSA generation for Neff analysis.** The number of effective sequences (Neff) was computed using MSAs generated via MMSeqs2 [51]. Toward this end, we searched all proteins in NEW364 against UniRef50 [41] using 3 iterations (−num_iterations 3). The UniRef50 hits were expanded by their UniRef100 cluster members (expandaln), effectively searching UniRef100 while reaching approximately the speed of searching UniRef50. The resulting MSA was clustered at 62% PIDE (filterresult −max-seq-id 0.62) to compute Neff.

**Pooling comparison.** When classifying a whole protein sequence instead of single tokens (protein-level predictions),

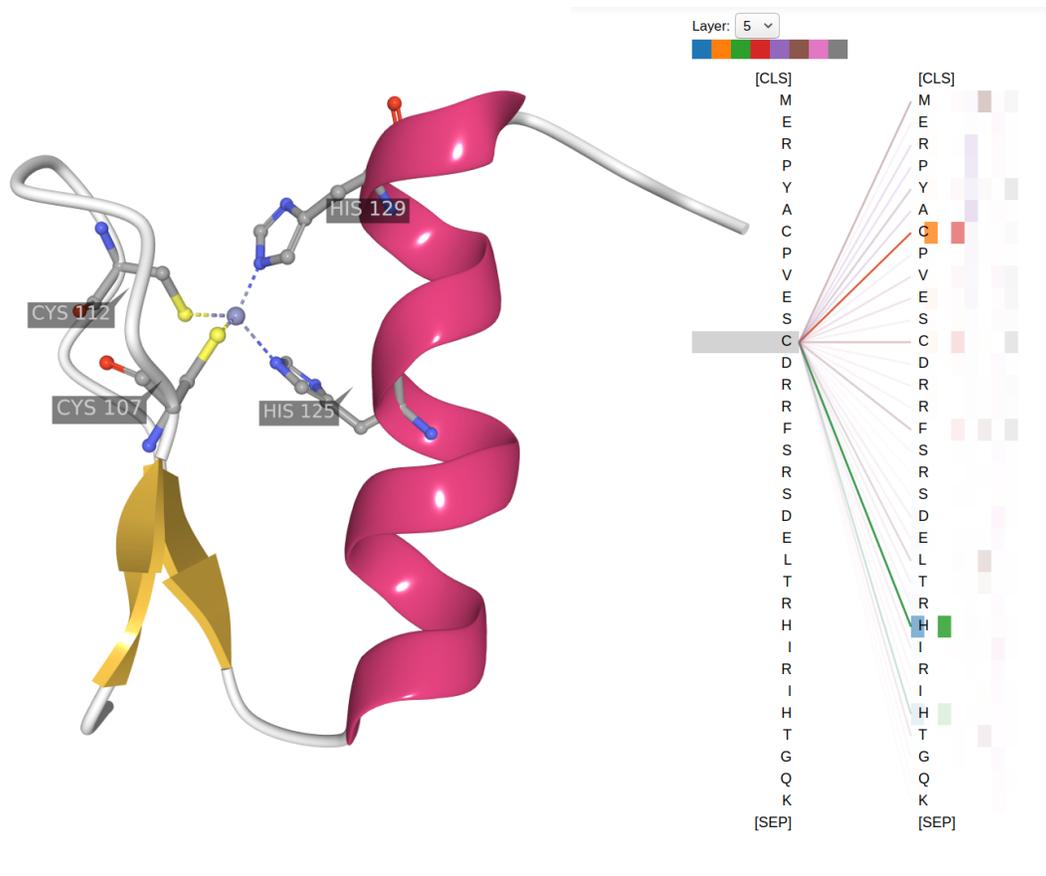

Fig. 11: Attention visualization - Here, we show how the inner workings of the Transformer's attention heads can be used to analyze single proteins in more detail. Each attention head performs internally an all-against-all comparison to compute weighted sums for each token over all other tokens in the sequence, i.e., in our case all residues in a protein are compared to each other. High scores indicate that the model learnt to put more weight on certain residue-pairs. Those scores can be indicative of a more fundamental biological truth, e.g. it was shown that some of these scores correlate with residue contacts [72]. Here, we have used bioviz [73] to visualize a protein structural motif that is crucial for DNA or RNA binding, i.e., the structure of the first 33 residues of a zinc-finger binding domain (PDB: 1A1L [101]) is shown on the left. The four residues that coordinate the zinc-binding in order to stabilize the fold are highlighted (C107, C112, H125, H129). On the right, we show a subset of the attention scores of one of our language models (ProtAlbert) for the same sequence. When visualizing the attention weights for C112, we observed that some of the attention heads in the fifth layer of ProtAlbert learnt to detect the Zinc-finger motif that is separate in sequence space but close in structure space. The attention weight is given by line thickness which indicates that the attention of residue C112 mostly attends to the three other residues (C107, H125, H129) involved in zinc-binding coordination. This shows exemplary how pre-trained Transformer models can be used for cheap and fast analysis that might open the door for novel ways of hypothesis generation.

one option is to pool token-level embeddings to derive a fixed-length protein representation from variable-length proteins. See Fig. 1 for an illustration of this process. Here, we compared four parameter-free pooling choices, i.e., min-, max-, mean-pooling (also called global average pooling) as well as a concatenation of those three. Classification accuracy when using those representations as an input for a FNN to differentiate between a) ten different subcellular localizations and b) membrane-bound and water soluble proteins is reported in Table 10. Min- and max-pooling perform significantly worse than mean-pooling. The concatenation of the three pooling strategies reaches the same performance as mean-pooling for differentiating between membrane-bound and water soluble proteins but falls significantly short on the classification of subcellular localization. One possible explanation for the latter observation is that the magnitudes of the concatenated vectors differed too much (no normaliziation was applied) while the dataset is too small for the network to learn to ignore the less informative inputs. Therefor, we stick to the mean-pooling strategy when comparing the language models trained here on classifying whole protein sequences.

### -1.4   Protein LM inference speed

In this section we compare the effect of sequence length on the time needed to extract features from the differen protein LMs trained here (SOM Fig. 13). Additionally, we analyse the cross-effect of sequence length and batch-size (SOM Table 11). Further, we add technical details to the human proteome benchmark shown in Fig. 9.

**Sequence length effect.** The effect of varying sequence lengths (128, 256, 512) and different batch sizes (1, 16, 32)

| Dataset | TS115 | CB513 |
|---|---|---|
| DeepProtVec | 66.5 | 63.7 |
| ProtTXL* | 75.3 | 73.7 |
| ProtTXL-BFD* | 75.3 | 73.5 |
| DeepSeqVec | 79.0 | 77.0 |
| ProtXLNet* | 80.5 | 77.9 |
| ProtElectra* | 81.2 | 79.2 |
| ProtAlbert* | 81.9 | 79.3 |
| ProtBert* | 82.9 | 80.9 |
| ProtBert-BFD* | 83.8 | 82.5 |
| ESM-1b | 84.8 | 83.9 |
| ProtT5-XXL-BFD* | 84.6 | 83.2 |
| ProtT5-XL-BFD* | 84.6 | 83.9 |
| ProtT5-XXL-U50* | 85.6 | 84.6 |
| ProtT5-XL-U50* | **86.9** | **86.2** |
| NetSurfP-2.0 | 85.7 | 85.4 |

TABLE 9: The three-state accuracy (Q3) for the per-residue/token-level secondary structure prediction (percentage of residues/tokens correctly predicted in either of three states: helix, strand, or other) for all protein language models (LMs) trained for this work (marked by star) along with other LMs, namely one word2vec-based approach (DeepProtVec), one LSTM (DeepSeqVec), one transformer (ESM-1b) and one of the current state-of-the-art methods (NetSurfP-2.0) that uses evolutionary information (EI)/multiple sequence alignments (MSAs). Values were compiled for two datasets: one because it is a standard in the field (CASP12, results for two other standard data sets - TS115 and CB513 - in Table 6), the other because it is larger and less redundant (dubbed NEW364 introduced here). Standard errors were computed using bootstrapping: CASP12=±1.6%, NEW364=±0.5%. Highest values in each column marked in bold-face.

| Pooling Strategy | Localization (Q10) | Membrane (Q2) |
|---|---|---|
| Min | 59 | 86 |
| Max | 60 | 85 |
| Mean | **74** | **89** |
| Concat | 64 | 89 |

TABLE 10: Pooling choice - One of our language models (ProtBERT-BFD) was used to compare different choices for pooling variable-length, token-level embeddings of a protein to a single, fixed-size protein-level representation, i.e., we compared the classification accuracy for differentiating 10 subcellular localizations and whether a protein is membrane-bound or water soluble. Global average pooling (Mean) outperforms min- or max-pooling and also the concatenation of all three pooling strategies (Concat) falls significantly short when classifying ten subcellular localizations while reaching the same performance for the classification of membrane-bound proteins.

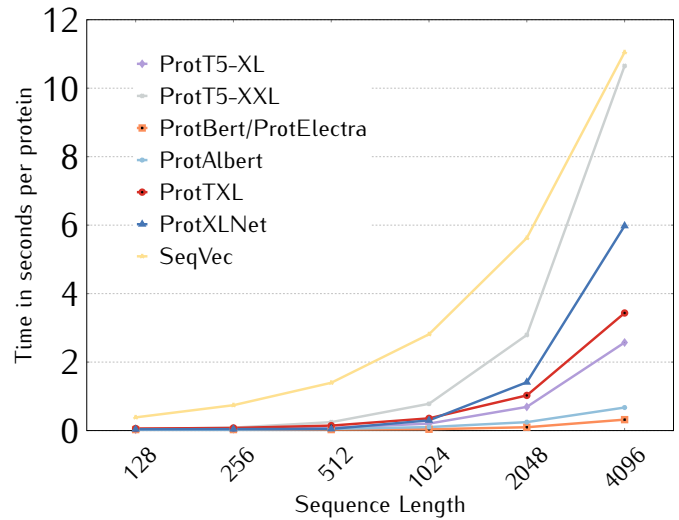

Fig. 13: Inference speed depends on sequence length: The effect of protein sequence length on the inference time of the protein LMs trained here and a previously published LM (SeqVec) were compared using a Nvidia Quadro RTX 8000 with 48GB memory using half precision (batch-size=1). Longer proteins take disproportionate long to embed for all language models. In particular, SeqVec was affected due to the sequential nature of the LSTMs used this LM, followed by ProtT5-XXL due to it is large number of parameters (5.5B for the encoder). In general, Transformer-based models require more inference time for longer for proteins because the attention maps that need to be computed square with sequence length. However, in contrast to LSTMs, the computation of attention can be parallelized, resulting in overall lower inference time for long proteins when using transformer-based LMs.

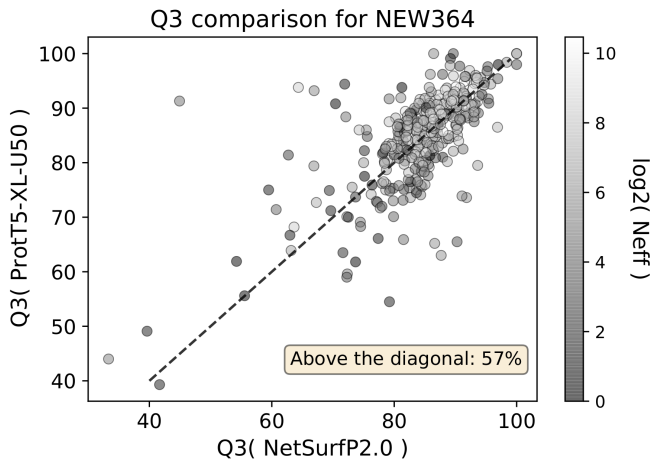

Fig. 12: Detailed Q3 comparison - We compared the 3-state secondary structure prediction performance (Q3) between one state-of-the-art method using evolutionary information (NetSurfP-2.0) and the best performing protein LM trained here (ProtT5-XL-U50) using our new test set (NEW364, see Methods). Dots resemble the Q3 achieved by each method for the same protein. Dots above the line indicate that the performance of our LM was better while dots below the diagonal show that NetSurfP-2.0 achieved higher performance. In total more than half of the proteins (57% or 208 out of 364) were predicted with higher accuracy by our method while using only single protein sequences.

on the inference time of the protein LMs trained here is reported in table 11. The effect of sequence length on different LM architectures (LSTM-based SeqVec and Transformer-based ProtTrans models) was visualized in figure 13. The x-axis refers to different sequence length from 128 up to 4096, while the y-axis represents the time of inference in ms for a single protein with a batch size of 1 on a Nvidia Quadro RTX8000 with 48GB.

### 1.5 Fast, proteome-wide feature-extraction.

We compared the time required to generate features between established methods (EI/MSA-based) and the proposed embeddings derived from the protein LMs trained here. Towards this end, we created embeddings for each protein in the human proteome (20,353) proteins with a median sequence length of

| Model | | ProtT5-XL | ProtT5-XXL | ProtBert | ProtAlbert | ProtElectra | ProtXLNet | ProtTXL | SeqVec |
|---|---|---|---|---|---|---|---|---|---|
| Sequence Length | Batch Size | | | | | | | | |
| 512 | 1 | 0.062 | 0.230 | 0.019 | 0.049 | 0.019 | 0.046 | 0.044 | 1.028 |
| | 16 | 0.054 | 0.227 | 0.013 | 0.044 | 0.13 | 0.098 | 0.054 | 0.078 |
| | 32 | 0.055 | 0.232 | 0.013 | 0.045 | 0.13 | 0.100 | 0.065 | 0.045 |
| 256 | 1 | 0.030 | 0.072 | 0.019 | 0.025 | 0.019 | 0.031 | 0.041 | 0.530 |
| | 16 | 0.015 | 0.062 | 0.005 | 0.021 | 0.005 | 0.023 | 0.012 | 0.039 |
| | 32 | 0.014 | 0.062 | 0.005 | 0.021 | 0.005 | 0.025 | 0.012 | 0.022 |
| 128 | 1 | 0.017 | 0.033 | 0.019 | 0.016 | 0.019 | 0.031 | 0.042 | 0.275 |
| | 16 | 0.006 | 0.023 | 0.003 | 0.10 | 0.003 | 0.006 | 0.004 | 0.021 |
| | 32 | 0.006 | 0.023 | 0.002 | 0.10 | 0.002 | 0.006 | 0.004 | 0.012 |
| Average | 1 | 0.036 | 0.112 | 0.019 | 0.03 | 0.019 | **0.036** | 0.042 | 0.611 |
| | 16 | 0.025 | **0.104** | 0.007 | 0.025 | 0.007 | 0.042 | **0.023** | 0.046 |
| | 32 | **0.025** | 0.106 | **0.007** | **0.025** | **0.007** | 0.044 | 0.027 | **0.026** |

TABLE 11: Comparison of inference speed: The analysis distinguished proteins of different length, as well as different batch sizes (numbers of proteins processed: 1, 16 and 32. For simplicity, no proteins longer than 512 is shown . Each test was repeated 100 times and the average time per protein was reported. The experiment was conducted using a single redNvidia Quadro RTX8000 with 48GB memory.

415 residues) using a) our protein LMs to generate embeddings and b) the fastest method available, namely MMseqs2 [51], to generate MSAs. In light of wanting to compare to the state-of-the-art secondary structure prediction method, NetSurfP-2.0 [15], we used the same parameters for the MMseqs2 search used by that method (*–num_iterations 2 –diff 2000*) and compared two databases (UniRef90 with 113M and UniRef100 with 216M proteins). All comparisons used an Intel® Xeon® Scalable Processor "Skylake" Gold 6248 with 40 threads, SSD and 377GB main memory, while protein LMs were computed on a single NVIDIA Quadro RTX 8000 with 48GB memory using half precision and dynamic batch size based on variable protein sequence lengths. MMseqs2 was about 16 to 28-times slower than the fastest LMs (ProtElectra and ProtBert), and about 4 to 6-times slower than our best model (ProtT5) (Fig. 9). The best performing model, ProtT5-XL-U50, required on average 0.12 seconds to create embeddings for a human protein, completing the entire human proteome (all proteins in an organism) in 40 minutes. We noticed that SeqVec was the slowest model for long proteins (11 seconds for protein with 4096 residues) while ProtBert was the fastest (0.32s) for those.

## -1.6   Additional Resources

For long-term storage the repository is also backed up through zenodo[9]. Additionally, all models are in the transformer library of *huggingface* [15][10]. Furthermore, secondary structure predictions for ProtBERT-BFD are available through the Predict-Protein [25] webserver[11]. Furthermore, the bio_embeddings [17] package[12] package could be used for simplified accessibility and analysis of protein LMs and predictions of the supervised models.

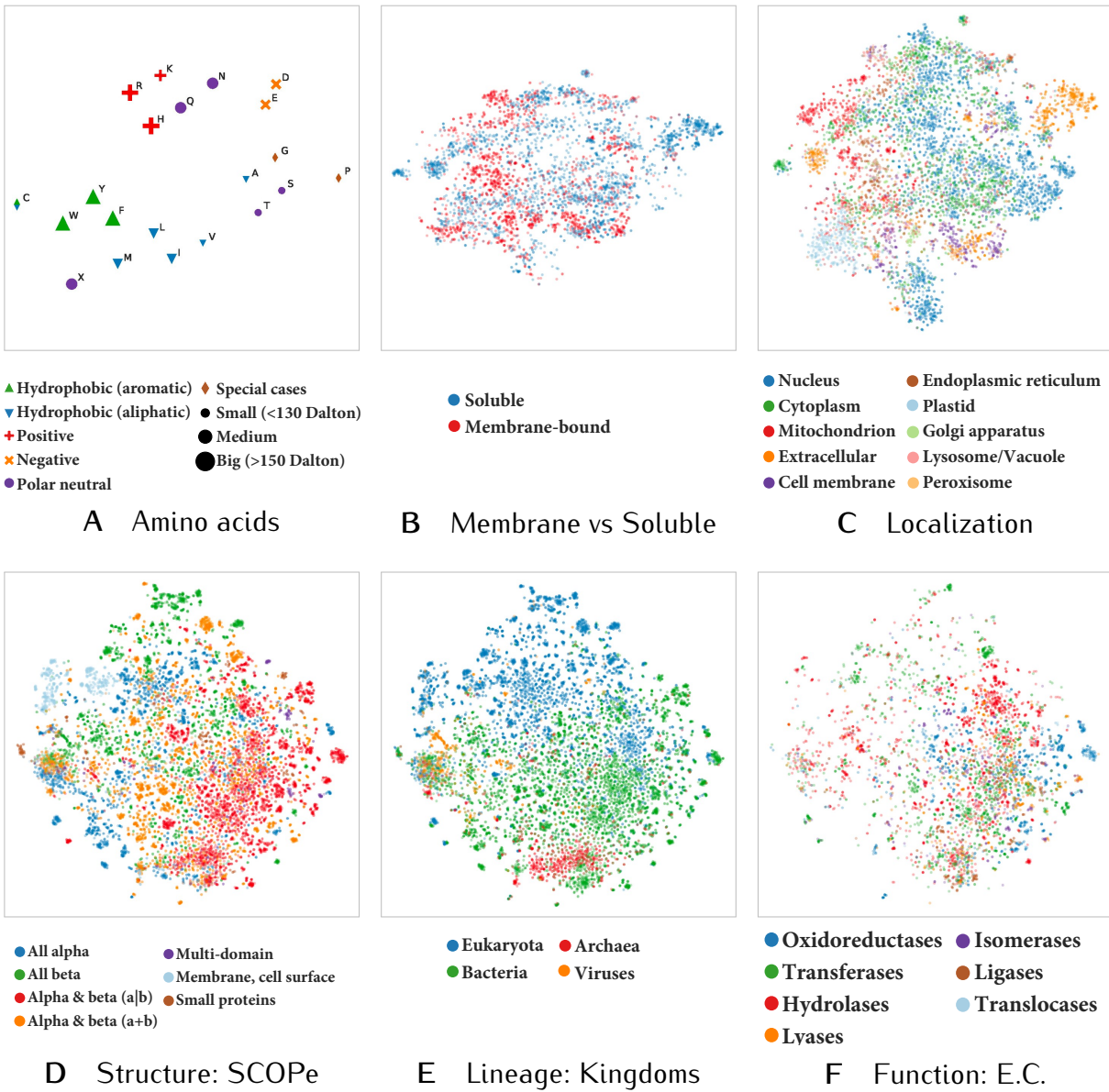

Fig. 14: Unsupervised training captures various features of proteins: We used t-SNE projections to assess which features the LMs trained here learnt to extract from proteins. Exemplarily for ProtT5-XL-U50, the best-performing model on supervised tasks, we showed that the protein LMs trained here captured biophysical- and biochemical properties of single amino acids during pre-training (Panel A). A redundancy reduced version (30%) of the DeepLoc [16] dataset was used to assess whether the LM learnt to classify proteins into membrane-bound and water-soluble (Panel B) or according to their cellular compartment (Panel C). Not all proteins in the set had annotations for both features, making Panels B and C not directly comparable. Further, a redundancy reduced version (40%) of the Structural Classification of Proteins – extended (SCOPe) database was used to assess whether ProtT5-XL captured structural (Panel D), functional (Panel F) or lineage-specific (Panel E) features of proteins without any labels. Towards this end, contextualized, fixed-size representations were generated for all proteins in both datasets by mean-pooling over the representations extracted from the last layer of ProtT5-XL (average over the length of the protein). The high-dimensional embeddings were projected to 2D using t-SNE. ProtT5-XL captured protein information on different levels: ranging from structural features as annotated in the main classes in SCOPe, over functional aspects as defined by in the Enzyme Commission (E.C.) numbers or the cellular compartment to the branch of the protein within the tree of life, without ever having been explicitly trained on any of these features. Comparing different features for the same datasets revealed that potentially heterogeneous clusters are only formed due to the multi-modal nature of proteins, e.g. the eukaryotic proteins are well separated from bacterial proteins (Panel E) but form internally multiple sub-clusters in structure space (Panel D).

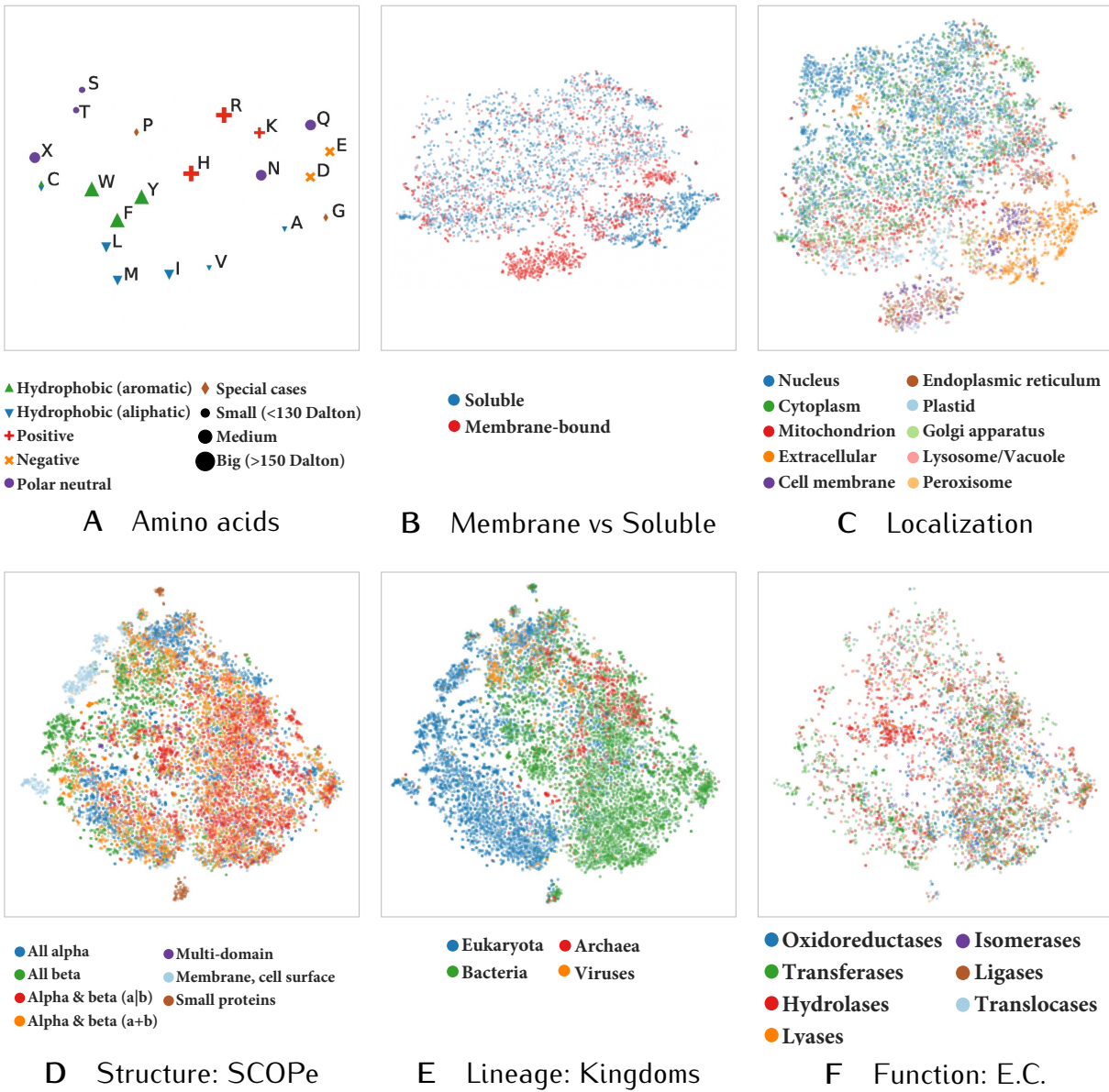

Fig. 15: Unsupervised training captures various features of proteins: We used t-SNE projections to assess which features the LMs trained here learnt to extract from proteins. Exemplarily for ProtBert-BFD, we showed that the protein LMs trained here captured biophysical- and biochemical properties of single amino acids during pre-training (Panel A). A redundancy reduced version (30%) of the DeepLoc [16] dataset was used to assess whether the LM learnt to classify proteins into membrane-bound and water-soluble (Panel B) or according to their cellular compartment (Panel C). Not all proteins in the set had annotations for both features, making Panels B and C not directly comparable. Further, a redundancy reduced version (40%) of the Structural Classification of Proteins – extended (SCOPe) database was used to assess whether ProtBert-BFD captured structural (Panel D), functional (Panel F) or lineage-specific (Panel E) features of proteins without any labels. Towards this end, contextualized, fixed-size representations were generated for all proteins in both datasets by mean-pooling over the representations extracted from the last layer of ProtBert-BFD (average over the length of the protein). The high-dimensional embeddings were projected to 2D using t-SNE. ProtBert-BFD captured protein information on different levels: ranging from structural features as annotated in the main classes in SCOPe, over functional aspects as defined by in the Enzyme Commission (E.C.) numbers or the cellular compartment to the branch of the protein within the tree of life, without ever having been explicitly trained on any of these features. Comparing different features for the same datasets revealed that potentially heterogeneous clusters are only formed due to the multi-modal nature of proteins, e.g. the eukaryotic proteins are well separated from bacterial proteins (Panel E) but form internally multiple sub-clusters in structure space (Panel D).

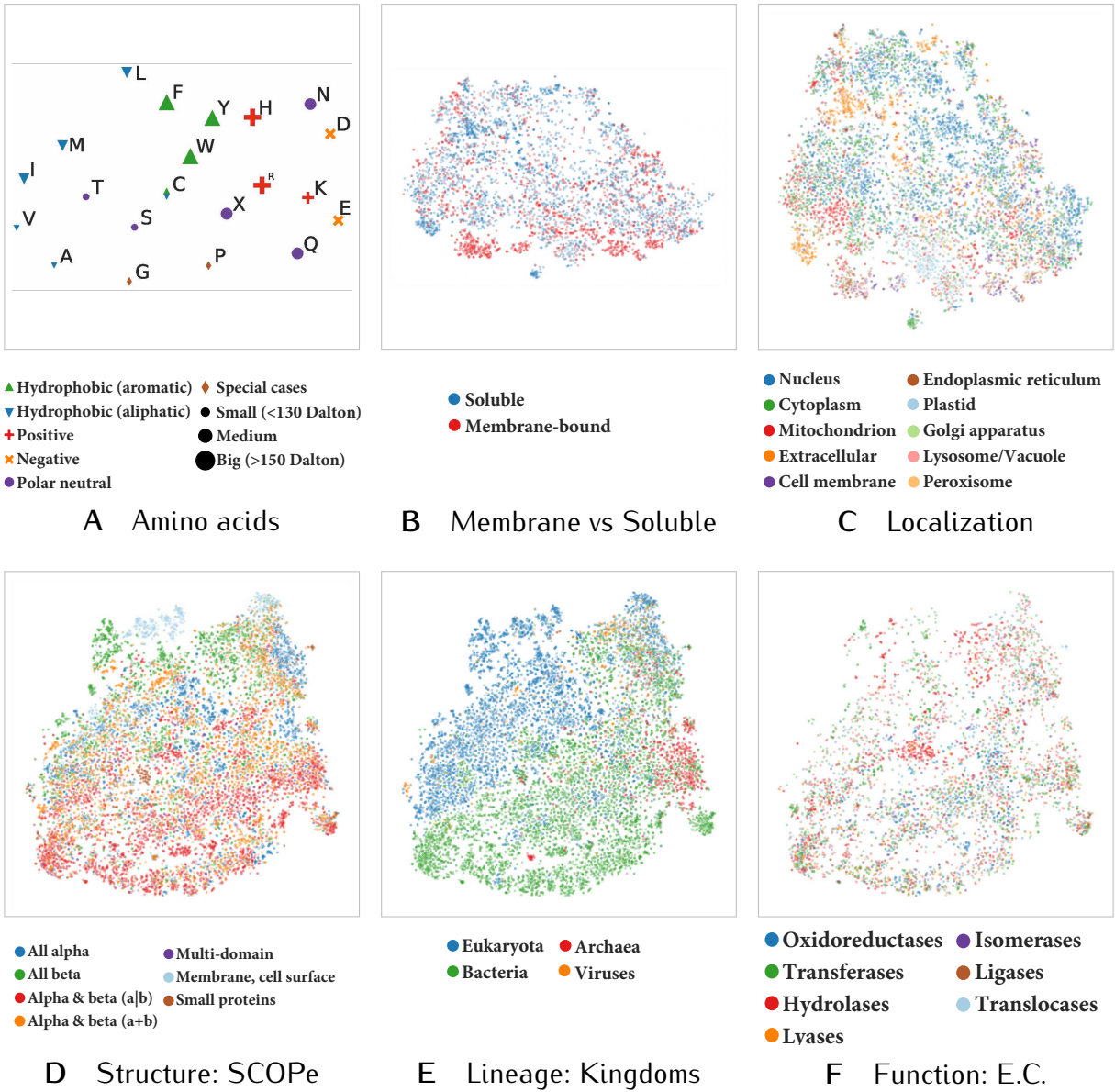

Fig. 16: Unsupervised training captures various features of proteins: We used t-SNE projections to assess which features the LMs trained here learnt to extract from proteins. Exemplarily for ProtBert, we showed that the protein language models trained here captured biophysical- and biochemical properties of single amino acids (Panel A). A redundancy reduced version (30%) of the DeepLoc ( [16]) dataset was used to assess whether ProtBert learnt to classify proteins into membrane-bound or water-soluble (Panel B) or according to the cellular compartment they appear in (Panel C). Not all proteins in the set had annotations for both features, making Panels B and C not directly comparable. Further, a redundancy reduced version (40%) of the Structural Classification of Proteins – extended (SCOPe) database was used to assess whether ProtBert captured structural (Panel D), functional (Panel F) or lineage-specific (Panel E) features of proteins without any labels. Towards this end, contextualized, fixed-size representations were generated for all proteins in both datasets by mean-pooling over the representations extracted from the last layer of ProtBert (average over the length of the protein). The high-dimensional embeddings were projected to 2D using t-SNE. ProtBert formed less dense clusters compared to the same model trained on a larger dataset (ProtBert-BFD Fig. 15).

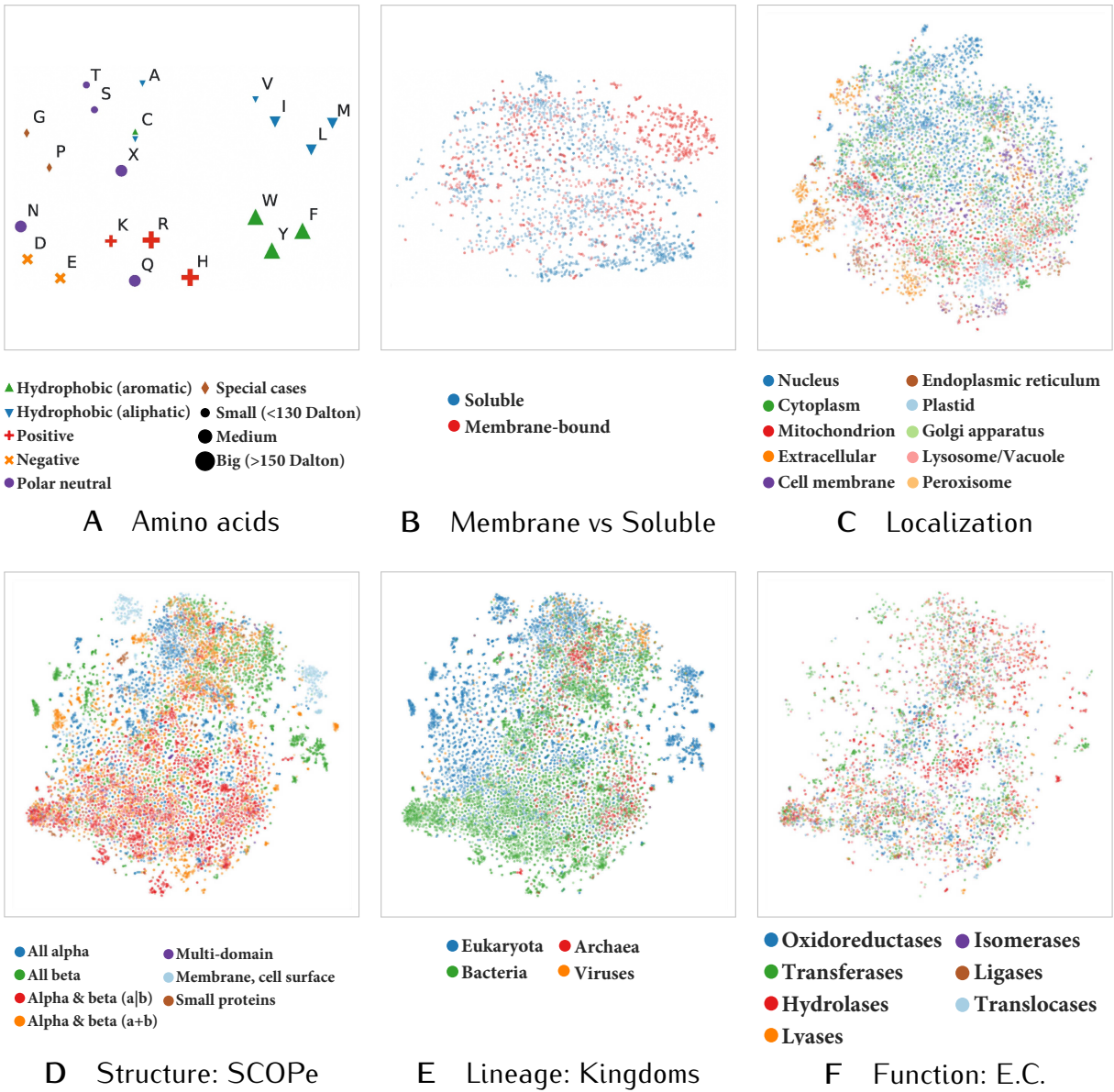

Fig. 17: Unsupervised training captures various features of proteins: We used t-SNE projections to assess which features the LMs trained here learnt to extract from proteins. Exemplarily for ProtAlbert, we showed that the protein language models trained here captured biophysical- and biochemical properties of single amino acids (Panel A). A redundancy reduced version (30%) of the DeepLoc ( [16]) dataset was used to assess whether ProtAlbert learnt to classify proteins into membrane-bound or water-soluble (Panel B) or according to the cellular compartment they appear in (Panel C). Not all proteins in the set had annotations for both features, making Panels B and C not directly comparable. Further, a redundancy reduced version (40%) of the Structural Classification of Proteins – extended (SCOPe) database was used to assess whether ProtAlbert captured structural (Panel D), functional (Panel F) or lineage-specific (Panel E) features of proteins without any labels. Towards this end, contextualized, fixed-size representations were generated for all proteins in both datasets by mean-pooling over the representations extracted from the last layer of ProtAlbert (average over the length of the protein). The high-dimensional embeddings were projected to 2D using t-SNE. Compared to the some of the other protein LMs trained here (ProtTXL 19 and ProtBert 16, ProtAlbert formed more dense clusters, especially for the projections based on the SCOPe dataset (Panels D, E and F).

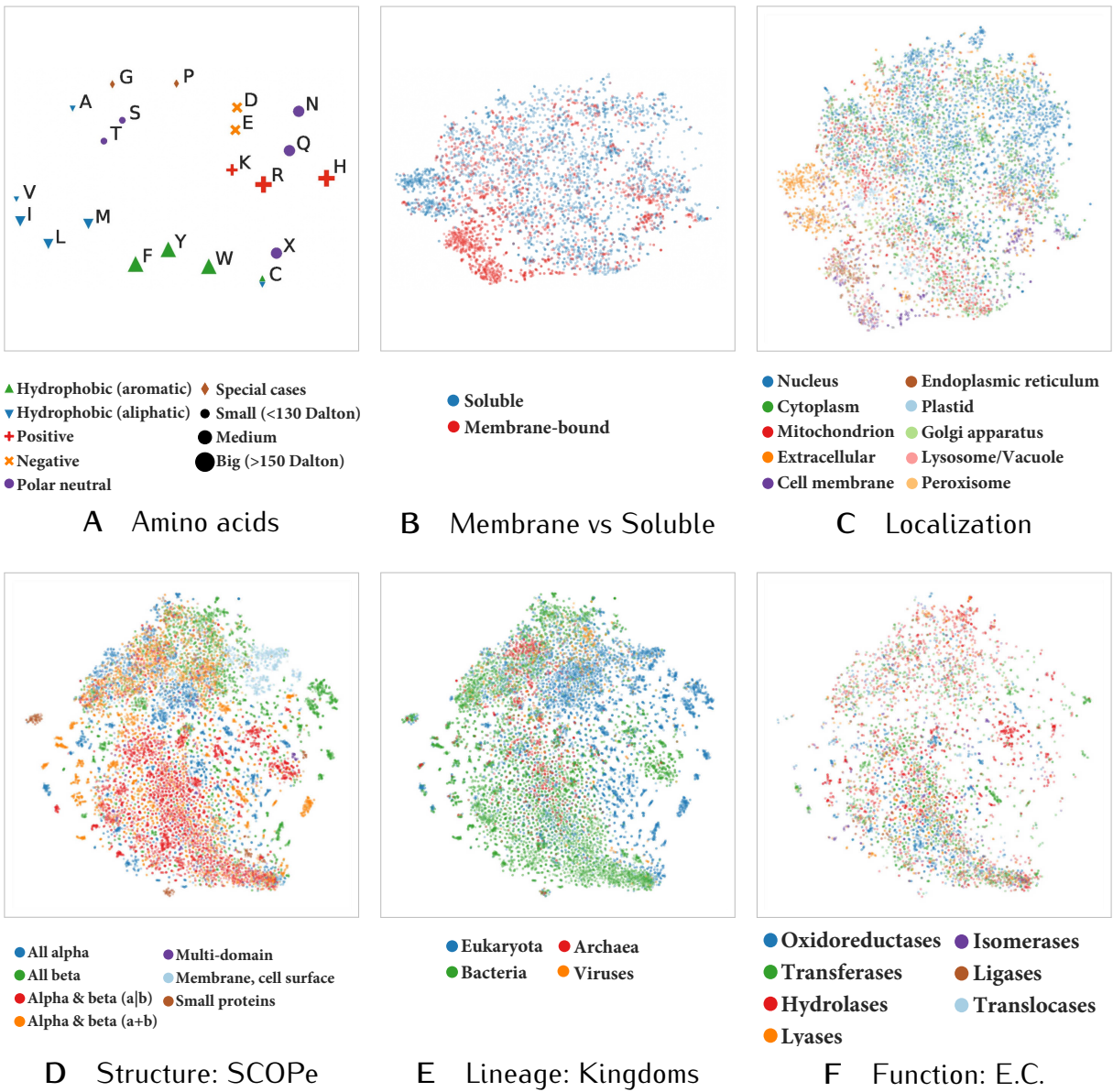

Fig. 18: Unsupervised training captures various features of proteins: We used t-SNE projections to assess which features the LMs trained here learnt to extract from proteins. Exemplarily for ProtXLNet, we showed that the protein language models trained here captured biophysical- and biochemical properties of single amino acids (Panel A). A redundancy reduced version (30%) of the DeepLoc ( [16]) dataset was used to assess whether ProtXLNet learnt to classify proteins into membrane-bound or water-soluble (Panel B) or according to the cellular compartment they appear in (Panel C). Not all proteins in the set had annotations for both features, making Panels B and C not directly comparable. Further, a redundancy reduced version (40%) of the Structural Classification of Proteins – extended (SCOPe) database was used to assess whether ProtXLNet captured structural (Panel D), functional (Panel F) or lineage-specific (Panel E) features of proteins without any labels. Towards this end, contextualized, fixed-size representations were generated for all proteins in both datasets by mean-pooling over the representations extracted from the last layer of ProtXLNet (average over the length of the protein). Compared to other protein LMs trained here, ProtXLNet learnt small coherent clusters that are scattered among the t-SNE projection. Only comparing different features for the same datasets reveals that potentially heterogenous clusters are only formed due to the mulit-modal nature of proteins, e.g. the eukaryotic proteins are well separated from bacterial proteins (Panel E but form multiple sub-clusters in structure space (Panel D). Compared to other protein LMs trained here (e.g. ProtTXL 19, ProtXLNet learnt small coherent clusters that are scattered among the t-SNE projection. Simlar to ProtBert-BFD, some of the small scattered clusters form homogeneous clusters when focusing on other aspects of proteins, e.g. some of the proteins in the heterogeneous cluster in the lower left part showing subcellular localization (Panel C) can be explained by proteins bound to the membrane (red in Panel B).

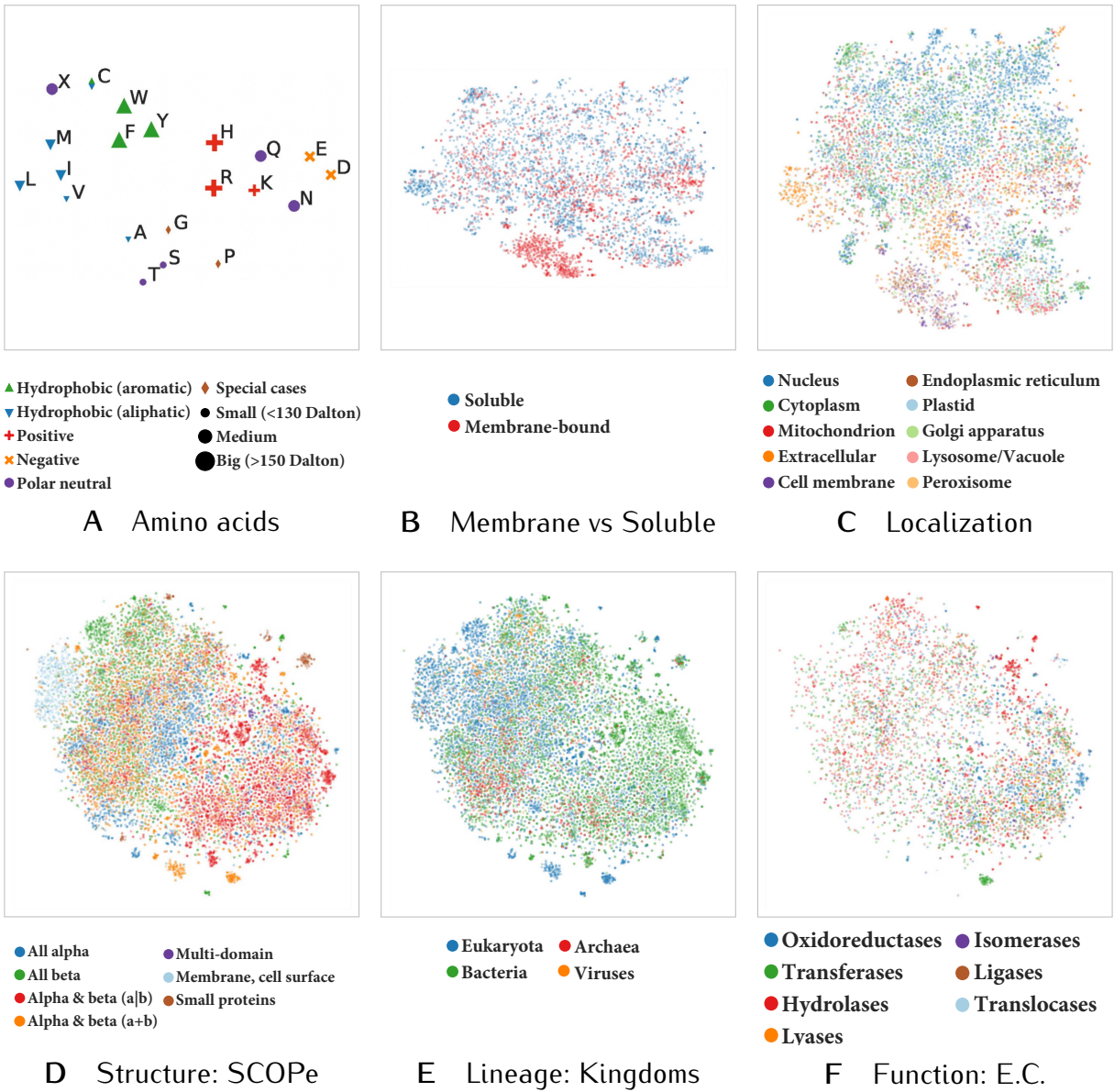

Fig. 19: Unsupervised training captures various features of proteins: We used t-SNE projections to assess which features the LMs trained here learnt to extract from proteins. Exemplarily for ProtTXL, we showed that the protein language models trained here captured biophysical- and biochemical properties of single amino acids (Panel A). A redundancy reduced version (30%) of the DeepLoc ( [16]) dataset was used to assess whether ProtTXL learnt to classify proteins into membrane-bound or water-soluble (Panel B) or according to the cellular compartment they appear in (Panel C). Not all proteins in the set had annotations for both features, making Panels B and C not directly comparable. Further, a redundancy reduced version (40%) of the Structural Classification of Proteins – extended (SCOPe) database was used to assess whether ProtTXL captured structural (Panel D), functional (Panel F) or lineage-specific (Panel E) features of proteins without any labels. Towards this end, contextualized, fixed-size representations were generated for all proteins in both datasets by mean-pooling over the representations extracted from the last layer of ProtTXL (average over the length of the protein). The high-dimensional embeddings were projected to 2D using t-SNE. While generally forming the least dense clusters compared to other LMs trained here, ProtTXL captured certain aspects about protein function (e.g. Transferases, dark green Panel F) that other LMs trained here did not capture.

# REFERENCES - SOM -